\documentclass[10pt,twocolumn,letterpaper]{article}

\usepackage{wacv}
\usepackage{times}
\usepackage{epsfig}
\usepackage{graphicx}
\usepackage{amsmath}
\usepackage{amssymb}

\usepackage[usenames]{color}
\usepackage{algorithm}
\usepackage{algcompatible}
\usepackage{epstopdf}
\usepackage{subcaption}
\usepackage[dvipsnames]{xcolor}
\usepackage[toc,page]{appendix}
\usepackage{algpseudocode}
\usepackage{arydshln}
\usepackage{multirow}

\usepackage[pagebackref=true,breaklinks=true,letterpaper=true,colorlinks,bookmarks=false]{hyperref}

\wacvfinalcopy 

\def\wacvPaperID{0035} 

\ifwacvfinal\fi
\begin{document}

\title{Saliency Driven Image Manipulation}

\author{Roey Mechrez\\
Technion\\
{\tt\small roey@tx.technion.ac.il}
\and
Eli Shechtman\\
Adobe Research\\
{\tt\small elishe@adobe.com}
\and
Lihi Zelnik-Manor\\
Technion\\
{\tt\small lihi@ee.technion.ac.il}
}

\maketitle
\thispagestyle{empty}

\newcommand*{\argmin}{arg\,min}

\newcommand*{\ShowNotes}{}

\definecolor{darkred}{rgb}{0.7,0.1,0.1}
\definecolor{darkgreen}{rgb}{0.1,0.7,0.1}
\definecolor{cyan}{rgb}{0.7,0.0,0.7}
\definecolor{dblue}{rgb}{0.2,0.2,0.8}
\definecolor{maroon}{rgb}{0.76,.13,.28}
\definecolor{burntorange}{rgb}{0.81,.33,0}

\ifdefined\ShowNotes
  \newcommand{\colornote}[3]{{\color{#1}\bf{#2: #3}\normalfont}}
\else
  \newcommand{\colornote}[3]{}
\fi

\newcommand {\note}[1]{\colornote{maroon}{Note}{#1}}
\newcommand {\todo}[1]{\colornote{cyan}{TODO}{#1}}
\newcommand {\lihi}[1]{\colornote{magenta}{LZ}{#1}}
\newcommand {\eli}[1]{\colornote{blue}{ES}{#1}}
\newcommand {\roey}[1]{\colornote{red}{RM}{#1}}

\newcommand{\ignorethis } [1] {}
\newcommand{\DB         }     {{\mathcal{D}}}
\newcommand{\THR        }     {{\tau}}
\newcommand{\shortcite       }     {{\cite}}

\begin{abstract}

Have you ever taken a picture only to find out that an unimportant background object ended up being overly salient? Or one of those team sports photos where your favorite player blends with the rest? Wouldn't it be nice if you could tweak these pictures just a little bit so that the distractor would be attenuated and your favorite player will stand-out among her peers? Manipulating images in order to control the saliency of objects is the goal of this paper. We propose an approach that considers the internal color and saliency properties of the image. It changes the saliency map via an optimization framework that relies on patch-based manipulation using only patches from within the same image to maintain its appearance characteristics. Comparing our method to previous ones shows significant improvement, both in the achieved saliency manipulation and in the realistic appearance of the resulting images.
\end{abstract}


\section{Introduction}

\begin{figure}[t]
		\centering
		\rotatebox{90}{\hspace{0.1cm}}~
		\vspace{0.1cm}
			\begin{subfigure}{.23\textwidth}
			\includegraphics[width=.98\linewidth]{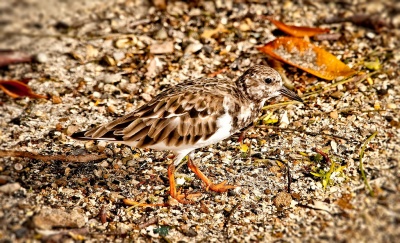}\caption{Input image}
			\label{fig:teaser-input}
			\end{subfigure}~
            \begin{subfigure}{.23\textwidth}
            \includegraphics[width=.98\linewidth]{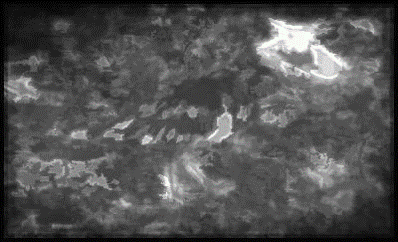}\caption{Input saliency map}
            \label{fig:teaser-input-saliency}
			\end{subfigure}
        \centering
		\rotatebox{90}{\hspace{0.1cm}}~
		\vspace{-0.1cm}
			\begin{subfigure}{.23\textwidth}
			\includegraphics[width=.98\linewidth]{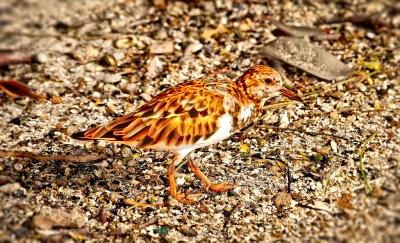}\caption{Manipulated image}
			\label{fig:teaser-output}
			\end{subfigure}~
			\begin{subfigure}{.23\textwidth}
			\includegraphics[width=.98\linewidth]{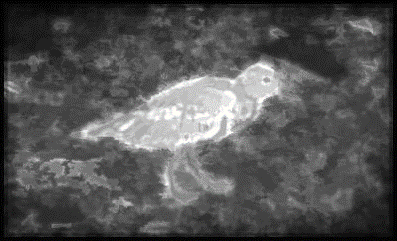}\caption{Manipulated saliency map}
			\label{fig:teaser-output-saliency}
			\end{subfigure}
\caption{Our saliency driven image manipulation algorithm can increase or decrease the saliency of a region. In this example the manipulation highlighted the bird while obscuring the leaf. This can be assessed both by viewing the image before (a) and after (c) manipulation, and by the corresponding saliency maps (b),(d) (computed using \cite{margolin2013makes}).}
\label{fig:teaser}
\end{figure}

Saliency detection, the task of identifying the salient and non-salient regions of an image, has drawn considerable amount of research in recent years, e.g., \cite{goferman2012context,li2013saliency,margolin2013makes,yan2013hierarchical,zhang2015minimum}. Our interest is in manipulating an image in order to modify its corresponding saliency map. This task has been named before as {\em attention retargeting} \cite{mateescu2014visual} or {\em re-attentionizing} \cite{nguyen2013image} and has not been explored much, even though it could be useful for various applications such as object enhancement \cite{mateescu2014attention,nguyen2013image}, directing viewer’s
attention in mixed reality~\cite{mendez2010focus} or in computer games~\cite{bernhard2011manipulating}, distractor removal \cite{fried2015finding}, background de-emphasis \cite{su2005emphasis} and improving image aesthetics \cite{hagiwara2011saliency,wong2011saliency,yan2015automatic}. Imagine being able to highlight your child who stands in the chorus line, or making it easier for a person with a visual impairment to find an object by making it more salient. Such manipulations are the aim of this paper.

Image editors use complex manipulations to enhance a particular object in a photo. They combine effects such as increasing the object's exposure, decreasing the background exposure, changing  hue, increasing saturation, or blurring the background. More importantly, they adapt the manipulation to each photo -- if the object is too dark they increase its exposure, if its colors are too flat they increase its saturation etc. Such complex manipulations are difficult for novice users that often do not know what to change and how. Instead, we provide the non-experts an intuitive way to highlight objects. All they need to do is mark the target region and tune a single parameter, that is directly linked to the desired saliency contrast between the target region and the rest of the image. An example manipulation is presented in Figure~\ref{fig:teaser}.

The approach we propose makes four key contributions over previous solutions. First, our approach handles multiple image regions and can either increase or decrease the saliency of each region. This is essential in many cases to achieve the desired enhancement effect. Second, we produce realistic and natural looking results by manipulating the image in a way that is consistent with its internal characteristics. This is different from many previous methods that enhance a region by recoloring it with a preeminent color that is often very non-realistic (e.g., turning leaves to cyan and goats to purple). Third, our approach provides the user with an intuitive way for controlling the level of enhancement. This important feature is completely missing from all previous methods.
Last, but not least, we present the first benchmark for object enhancement that consists of over 650 images. This is at least an order of magnitude larger than the test-sets of previous works, that were satisfied with testing on a very small number of cherry-picked images.

The algorithm we propose aims at globally optimizing an overall objective that considers the image saliency map. A key component to our solution is replacing properties of image patches in the target regions with other patches from the same image. This concept is a key ingredient in many patch-bases synthesis and analysis methods, such as texture synthesis~\cite{efros1999texture},  image completion~\cite{barnes2009patchmatch},  highlighting irregularities \cite{boiman2007detecting}, image summarization~\cite{simakov2008summarizing}, image compositing and harmonization~\cite{DarabiImageMelding12} and recently highlighting non-local variations~\cite{dekel2015revealing}. Our method follows this line of work as we replace patches in the target regions with similar ones from other image regions. Differently from those methods, our patch-to-patch similarity considers the saliency of the patches with respect to the rest of the image. This is necessary to optimize the saliency-based objective we propose. A key observation we make is that these patch replacements do not merely copy the saliency of the source patch to the target location as saliency is a complex global phenomena (similar idea was suggested in \cite{cheng2015global} for saliency detection). Instead, we interleave saliency estimation within the patch synthesis process. In addition, we do not limit the editing to the target region but rather change (if necessary) the entire image to obtain the desired global saliency goal.

we propose a new quantitative criteria to assess performance of saliency editing algorithms
by comparing two properties to previous methods: (i) The ability to manipulate an image such that the saliency map of the result matches the user goal. (ii) The realism of the manipulated image. These properties are evaluated via qualitative means, quantitative measures and user studies. Our experiments show a significant improvement over previous methods. We further show that our general framework is applicable to two other applications: distractor attenuation and background decluttering.



\section{Related Work}
\label{sec:related}

Attention retargeting methods have a mutual goal -- to enhance a selected region. They differ, however, in the way the image is manipulated \cite{hagiwara2011saliency,mateescu2014attention,nguyen2013image,su2005emphasis,wong2011saliency}. We next briefly describe the key ideas behind these methods. A more thorough review and comparison is provided in~\cite{mateescu2014visual}. 

Some approaches are based solely on color manipulation~\cite{mateescu2014attention,nguyen2013image}. This usually suffices to enhance the object of interest, but often results in non-realistic manipulations, such as purple snakes or blue flamingos. Approaches that integrate also other saliency cues, such as saturation, illumination and sharpness have also been proposed~\cite{hagiwara2011saliency,mendez2010focus,su2005emphasis,wong2011saliency}. While attempting to produce realistic and aesthetic results, they do not always succeed, as we show empirically later on.

Recently Yan et al.~\cite{yan2015automatic} suggested a deep convolutional network to learn transformations that adjust image aesthetics.  One of the effects they study is Foreground Pop-Out, which is similar in spirit to object saliency enhancement. Their method produces aesthetic results, however, it requires intensive manual labeling by professional artists in the training phase and it is limited to the labeled effect used by the professional. 

\section{Problem Formulation}
\label{sec:problem}

Our \emph{Object Enhancement} formulation takes as input an image $I$, a target region mask $R$
and the desired saliency contrast $\Delta S$ between the target region and the rest of the image. 
It generates a manipulated image $J$ whose corresponding saliency map is denoted by $S_J$.

We pose this task as a patch-based optimization problem over the image $J$. The objective we define distinguishes between salient and non-salient patches and pushes for manipulation that matches the saliency contrast $\Delta S$.
To do this we extract from the input image $I$ two databases of patches of size $w\!{\times}\! w$: $\DB^{+} = \{p; S_I(p) \geq \mathcal{\THR}^{+} \}$ of patches $p$ with {\em high} saliency and $\DB^{-}  = \{p; S_I(p) \leq \mathcal{\THR}^{-} \}$ of patches $p$ with {\em low} saliency. The thresholds $\mathcal{\THR}^{+}$ and $\mathcal{\THR}^{-}$ are found via our optimization (explained below).

To increase the saliency of patches $\in\!{R}$ and decrease the saliency of patches $\notin\!{R}$ we define the following energy function:
\begin{eqnarray}
\label{eq:energy}
E(J,\DB^+,\DB^-) & = & E^+ + E^- + \lambda \cdot E^{\nabla} \\
E^+(J,\DB^+) & = & \sum_{q \in R}{\min_{p \in \DB^{+}}{D(q,p)}} \nonumber \\
E^-(J,\DB^-) & = & \sum_{q \notin R}{\min_{p \in \DB^{-}}{D(q,p)}} \nonumber\\
E^{\nabla}(J,I) & = & \|\nabla J - \nabla I \|_2 \nonumber
\end{eqnarray}
where $D(q,p)$ is the sum of squared distances (SSD) over $\{L,a,b\}$ color channels between patches $q$ and $p$.
The role of the third term, $E^{\nabla}$, is to preserve the gradients of the original image $I$. The balance between the color channels and the gradient channels is controlled by $\lambda$.

Recall, that our goal in minimizing ~\eqref{eq:energy} is to generate an image $J$ with saliency map $S_J$, such that the contrast in saliency between $R$ and the rest of the image is $\Delta S$. The key to this lies in the construction of the patch sets $\DB^+$ and $\DB^-$. The higher the threshold $\mathcal{\THR}^{+}$ the more salient will be the patches in $\DB^+$ and in return those in $R$. Similarly, the lower the threshold $\mathcal{\THR}^{-}$ the less salient will be the patches in $\DB^-$ and in return those outside of $R$. Our algorithm performs an approximate greedy search over the thresholds to determine their values.

To formulate mathematically the affect of the user control parameter $\Delta S$ we further define a function $\psi(S_J,R)$ that computes the saliency difference between pixels in the target region $R$ and those outside it:
\begin{equation}
\label{eq:contrast-function}
\psi(S_J,R) = \underset{\beta_{top}}{\mathrm{mean}} \{S_J \in R\} - \underset{\beta_{top}}{\mathrm{mean}} \{S_J \notin R\}
\end{equation}
and seek to minimize the saliency-based energy term: 
\begin{equation}
\label{eq:saliency-energy}
E^{sal} = \| \psi(S_J,R) - \Delta S \|
\end{equation}
For robustness to outliers we only consider the $\beta_{top}$ ($ = 20\%$) most salient pixels in $R$ and outside $R$ in the mean calculation.


\section{Algorithm Overview}
\label{sec:overview}

The optimization problem in~\eqref{eq:energy} is non-convex with respect to the databases $\DB^+$, $\DB^-$. To solve it, we perform an approximate greedy search over the thresholds $\mathcal{\THR}^{+}$, $\mathcal{\THR}^{-}$  to determine their values. Given a choice of threshold values, we construct the corresponding databases and then minimize the objective in~\eqref{eq:energy} w.r.t. $J$, while keeping the databases fixed. 
Pseudo-code is provided in Algorithm~\ref{alg1}.

\begin{algorithm}[tb]
   \caption{Saliency Manipulation}
     \label{alg1}
\begin{algorithmic}[1]
\State{\textbf{Input}: Image $I$; object mask $R$; saliency contrast $\Delta S$.} 
\State{\textbf{Output}: Manipulated image $J$.}
\Statex{}
\State{Initialize $\mathcal{\THR}^{+}$, $\mathcal{\THR}^{-}$ and $J = I$.}
	\While {$\|\psi(S_J,R)-\Delta S\|>\epsilon$ *}
		\State {1. \textbf{Database Update}}
			\State  {\quad  $\rightarrow$ Increase $\mathcal{\THR}^{+}$ and decrease $\mathcal{\THR}^{-}$.}
		\State {2. \textbf{Image Update}}
			\State  {\quad  $\rightarrow$ Minimize \eqref{eq:energy} w.r.t. $J$, holding $\DB^+$,$\DB^-$ fixed.}
	\EndWhile
	\State {\textbf{Fine-scale Refinement}}
    \Statex {* the iterations also stopped when the $\mathcal{\THR}^{+}$ and $\mathcal{\THR}^{-}$ stop changing between subsequent iterations.}
\end{algorithmic}
\end{algorithm}

\textbf{Image Update:} 
Manipulate $J$ to enhance the region $R$. Patches $\in{R}$ are replaced with similar ones from $\DB^{+}$, while, patches $\notin{R}$ are replaced with similar ones from $\DB^{-}$.

\textbf{Database Update:} 
Reassign the patches from the input image $I$ into two databases, $\DB^{+}$ and $\DB^{-}$, of salient and non-salient patches, respectively. The databases are updated at every iteration by shifting the thresholds $\mathcal{\THR}^{+},\mathcal{\THR}^{-}$, in order to find values that yield the desired foreground enhancement and background demotion effects (according to $\Delta S$).

\textbf{Fine-scale Refinement:}
We observed that updating both the image $J$ and the databases $\DB^+,\DB^-$, at all scales, does not contribute much to the results, as most changes happen already at coarse scales. Similar behavior was observed by \cite{simakov2008summarizing} in retargeting and by \cite{barnes2009patchmatch} in reshuffling. 
Hence, the iterations of updating the image and databases are performed only at coarse resolution. After convergence, we continue and apply the Image Update step at finer scales, while the databases are held fixed. Between scales, we down-sample the input image $I$ to be of the same size as $J$, and then reassign the patches from the scaled $I$ into $\DB^{+}$ and $\DB^{-}$ using the current thresholds.

In our implementation we use a Gaussian pyramid with 0.5 scale gaps, and apply 5-20 iterations, more at coarse scales and less at fine scales. The coarsest scale is set to be 150 pixels width.

\section{Detailed Description of the Algorithm}
\label{sec:detail}


\paragraph{Saliency Model}
Throughout the algorithm when a saliency map is computed for either $I$ or $J$ we use a modification of~\cite{margolin2013makes}. Because we want the saliency map to be as sharp as possible, we use a small patch size of $5\!\times\! 5$. In addition, we omit the center prior which assumes higher saliency for patches at the center of the image. We found it to ambiguate the differences in saliency between patches, which might be good when comparing prediction results to smoothed ground-truth maps, but not for our purposes.
We selected the saliency estimation of~\cite{margolin2013makes} since its core is to find what makes a patch distinct. It assigns a score${\in}[0,1]$ to each patch based on the inner statistics of the patches in the image, which is a beneficial property to our method.

\begin{figure}[t]
		\centering
		\rotatebox{90}{\hspace{0.1cm}}~
		\vspace{0.1cm}
		\begin{subfigure}[b]{.2\textwidth}\includegraphics[width=.98\linewidth]{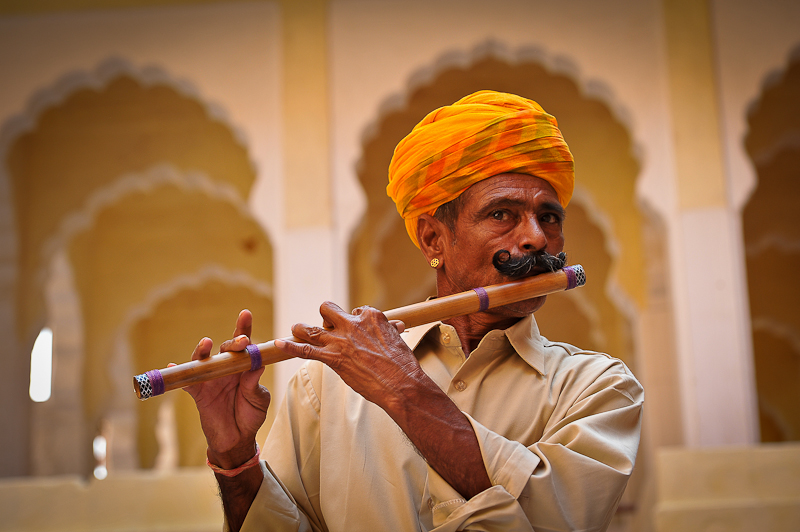}\caption{Input image $I$}\end{subfigure}
		\begin{subfigure}[b]{.2\textwidth}\includegraphics[width=.98\linewidth]{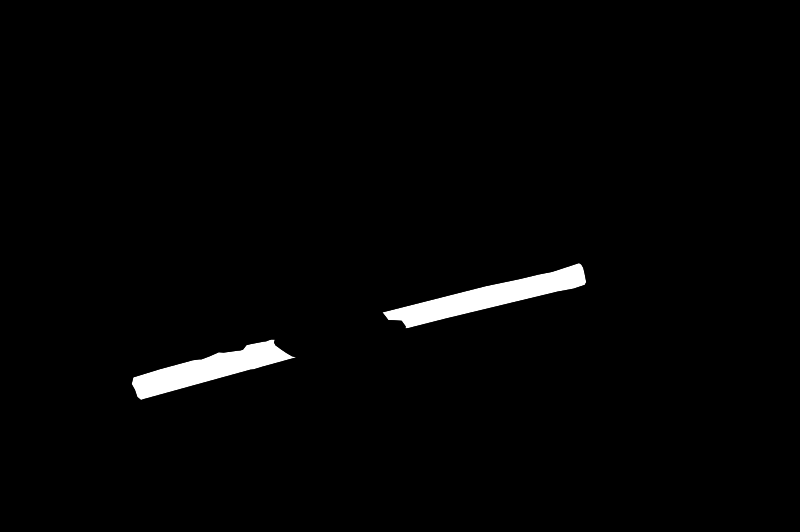}\caption{Mask $R$}\end{subfigure}
		\rotatebox{90}{\hspace{0.1cm}}~
		\vspace{-0.2cm}
		\begin{subfigure}[b]{.2\textwidth}\includegraphics[width=.98\linewidth]{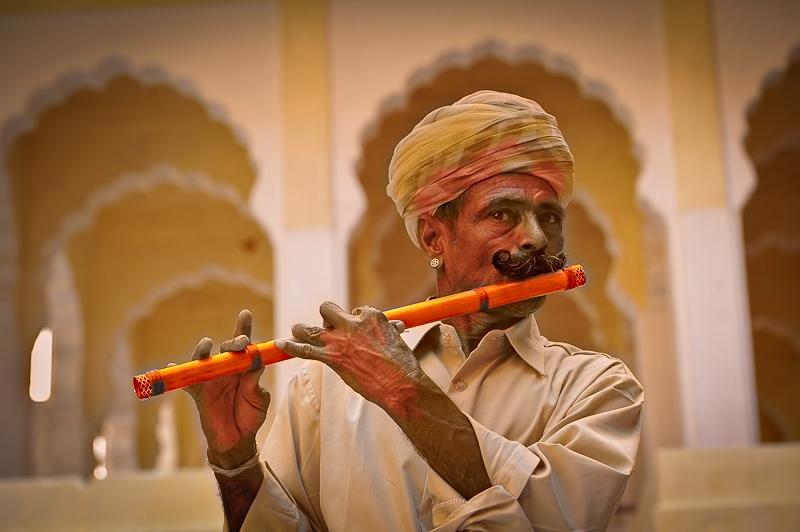}\caption{Without color gradients}\label{fig:color_grad_c}\end{subfigure}~
		\begin{subfigure}[b]{.2\textwidth}\includegraphics[width=.98\linewidth]{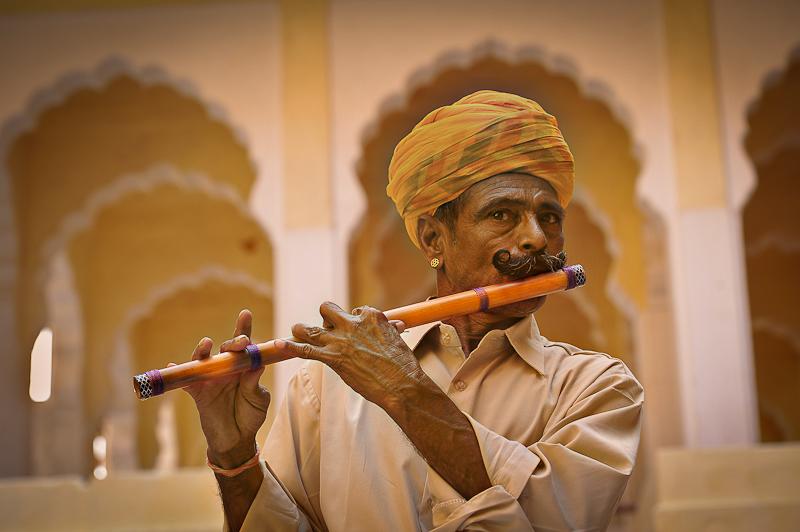}\caption{With color gradients}\label{fig:color_grad_d}\end{subfigure}
		\caption{\textbf{Chromatic gradients.}  A demonstration of the importance of chromatic gradients. (c) When \emph{not} using color gradients - artifacts appear: orange regions on the flutist' hat, hands and face. (d) By solving the screened Poisson equation on all three channels we improve the smoothness of the color manipulation, stopping it from generating spurious color edges, and the color of the flute is more natural looking.}
		\label{fig:color_grad}
\end{figure}

\vspace{-0.3cm}
\paragraph{Image Update}
In this step we minimize~\eqref{eq:energy} with respect to $J$, while holding the databases fixed.
This resembles the optimization proposed by~\cite{DarabiImageMelding12} for image synthesis.
It differs, however, in two important ways.
First, ~\cite{DarabiImageMelding12} consider only luminance gradients, while we consider gradients of all three $\{L,a,b\}$ color channels. This improves the smoothness of the color manipulation, preventing generation of spurious color edges, like those evident in Figure~\ref{fig:color_grad_c}.
It guides the optimization to abide to the color gradients of the original image and often leads to improved results (Figure~\ref{fig:color_grad_d}).

\ignorethis{A second difference is that our energy function has two components, one for the target region and another for the rest of the image.
To address this we split the problem into two: increase saliency ($E^+$) and decrease saliency ($E^-$). For each of these, we search for Nearest Neighbor patches separately from two different sources ($\DB^{+},\DB^{-}$), and then perform one voting step over all NN patches.}

As was shown in~\cite{DarabiImageMelding12}, the energy terms in~\eqref{eq:energy} can be optimized by combining a patch {\em search-and-vote} scheme and a discrete Screened Poisson equation that was originally suggested by~\cite{bhat2008fourier} for gradient domain problems. At each scale, every iteration starts with a {\em search-and-vote} scheme that replaces patches of color with similar ones from the appropriate patch database. For each patch $q\!\in\!J$ we search for the Nearest Neighbor patch $p$. Note, that we perform two separate searches, for the target region in $\DB^{+}$ and for the background in $\DB^{-}$. This is the second difference from~\cite{DarabiImageMelding12} where a single search is performed over one source region.

To reduce computation time the databases are represented as two images: $I_{\DB^+} = I \cap (S_I \geq \THR^+)$ and
$I_{\DB^-} = I \cap (S_I \leq \THR^-)$. The search is performed using PatchMatch~\cite{barnes2009patchmatch} with patch size $7 \! \times \! 7$ and translation transformation only (we found that rotation and scale were not beneficial).
In the {\em vote} step, every target pixel is assigned the mean color of all the patches that overlap with it. The voted color image is then combined with the original gradients of image $I$ using a Screened Poisson solver to obtain the final colors of that iteration. We fixed $\lambda = 5$ as the gradients weight.

Having constructed a new image $J$, we compute its saliency map $S_J$ to be used in the database update step explained next.

\vspace{-0.3cm}
\paragraph{Database Update}
The purpose of the database update step is to search for the appropriate thresholds that split the patches of $I$ into salient $\DB^{+}$ and non-salient $\DB^{-}$ databases.
Our underlying assumption is that there exist threshold values that  result in minimizing the objective $E^{sal}$ of~\eqref{eq:saliency-energy}.

Recall that the databases are constructed using two thresholds on the saliency map $S_I$ such that $\DB^{+} = \{p; S_I(p) \geq \mathcal{\THR}^{+} \}$ and $\DB^{-}  = \{p; S_I(p) \leq \mathcal{\THR}^{-} \}$.
An exhaustive search over all possible threshold values is non-tractable.
Instead, we perform an approximate search that starts from a low value for $\mathcal{\THR}^{+}$ and a high value for $\mathcal{\THR}^{-}$ and then gradually increases the first and reduces the second until satisfactory values are found. Note, that $\DB^{+}$ and $\DB^{-}$ could be overlapping if $\mathcal{\THR}^{+} < \mathcal{\THR}^{-}$.

The naive thresholds $\mathcal{\THR}^{+}\approx 1, \mathcal{\THR}^{-}\approx 0$, would leave only the most salient patches in $\DB^{+}$ and the most non-salient in $\DB^{-}$.
This, however, could lead to non-realistic results and might not match the user's input for a specific saliency contrast $\Delta S$.
To find a solution which considers realism and the user's input we seek the maximal $\mathcal{\THR}^{-}$ and minimal $\mathcal{\THR}^{+}$ that minimize the saliency term $E^{sal}$.


At each iteration we continue the search over the thresholds by gradually updating them:
\begin{eqnarray}
\mathcal{\THR}^{+}_{n+1} & = & \mathcal{\THR}^{+}_{n} + \eta \cdot \|\psi(S_J,R)-\Delta S\|\\
\mathcal{\THR}^{-}_{n+1} & = & \mathcal{\THR}^{-}_{n} - \eta \cdot \|\psi(S_J,\overline{R})-\Delta S\|
\end{eqnarray}
where $\overline{R}$ is the inverse of the target region $R$.
Since the values of the thresholds are not bounded, we trim them to be in the range of $[0,1]$.
Convergence is declared when $E^{sal}=\|\psi-\Delta S\|<\epsilon$, i.e.,  when the desired contrast is reached. If convergence fails the iterations are stopped when the thresholds stop changing between subsequent iterations. In our implementation $\eta=0.1$ and $\epsilon=0.05$.


An important property of our method is that if $\mathcal{\THR}^{-}= 1$ (or very high) and $\mathcal{\THR}^{+} = 0$ (or very low) the image would be left unchanged as the solution where all patches are replaced by themselves will lead to a zero error of our objective energy function~\eqref{eq:energy}.




\vspace{-0.3cm}
\paragraph{Robustness to parameters}
The only parameter we request the user to provide is $\Delta S$ which determines the enhancement level. We argue that this parameter is easy and intuitive to tune as it directly relates to the desired saliency contrast between the target region and the background.
We used a default value of $\Delta S=0.6$, for which convergence was achieved for $95\%$ of the images.
In only a few cases the result was not aesthetically pleasing and we used other values in the range $[0.4,0.8]$. Throughout the paper, if not mentioned otherwise, $\Delta S=0.6$.

An additional parameter is $\lambda$, which was fixed to $\lambda=5$ in our implementation. In practice, we found that for any value $\lambda>1$ we got approximately the same results, while for $\lambda<1$ the manipulated images tend to be blurry (mathematical analysis can be found in~\cite{bhat2008fourier}, since our $\lambda$ is equivalent to that of the screened Poisson).

\vspace{-0.3cm}
\paragraph{Convergence and speed}
Our algorithm is not guaranteed to reach a global minima. However we found that typically the manipulated image is visually plausible, and pertains a good match to the desired saliency.

It takes around 2 minutes to run our algorithm on a $1000~\times~1000$ image -- the most time demanding step of our method is solving the screened Poisson equation at each iteration. Since our main focus was on quality we did not optimize the implementation for speed. Significant speed-up could be achieved by adopting the method of \cite{farbman2011convolution}. As was shown by \cite{DarabiImageMelding12} replacing these fast pyramidal convolutions with our current solver, will reduce run-time from minutes to several seconds.

\section{Empirical Evaluation}
\label{sec:evaluation}

To evaluate object enhancement one must consider two properties of the manipulated image: (i) the similarity of its saliency map to the user-provided target, and, (ii) whether it looks realistic. Through these two properties we compare our algorithm to HAG~\cite{hagiwara2011saliency},  OHA~\cite{mateescu2014attention}, and
WSR~\cite{wong2011saliency}, that were identified as top performers in~\cite{mateescu2014visual}.
\footnote{Code for WSR and HAG is not publicly available, hence we used our own implementation that led to similar results on examples from their papers. This code publicly available for future comparisons in our webpage. For OHA we used the original code. }. 


We start by providing a qualitative sense of what our algorithm can achieve in Figure~\ref{fig:enhance}. Many more results are provided in the supplementary, and we encourage the reader to view them.
Comparing to OHA, it is evident that our results are more realistic. OHA changes the hue of the selected object such that its new color is unique with respect to the color histogram of the rest of the image. This often results in unrealistic colors.
The results of WSR and HAG, on the other hand, are typically realistic since their manipulation is restricted not to deviate too much from the original image in order to achieve realistic outcomes. This, however, comes at the expense of often failing to achieve the desired object enhancement altogether.

The ability of our approach to simultaneously reduce and increase saliency of different regions is essential in some cases, e.g. Figure~\ref{fig:enhance}, rows 1 and 4. In addition, it is important to note that our manipulation latches onto the internal statistics of the image and emphasizes the objects via a combination of different saliency cues, such as color, saturation  and illumination. Examples of these complex effects are presented in Figure~\ref{fig:enhance}, rows 2, 6 and 7, respectively.


\textbf{A new benchmark:} To perform quantitative evaluation we built a corpus of $667$ images gathered from previous papers on object enhancement and saliency \cite{bell2014trinsic,Cheng_Saliency,fried2015finding,Judd_2009,liu2010tud,mateescu2014attention} as well as images from MS COCO~\cite{lin2014microsoft}.
Our dataset is the largest ever built and tested for this task and sets a new benchmark in this area. Our dataset, code and results are publicly available \footnote{\url{http://cgm.technion.ac.il/people/Roey/}}.

\begin{figure}[t]
	\centering
\includegraphics[width=.9\linewidth]{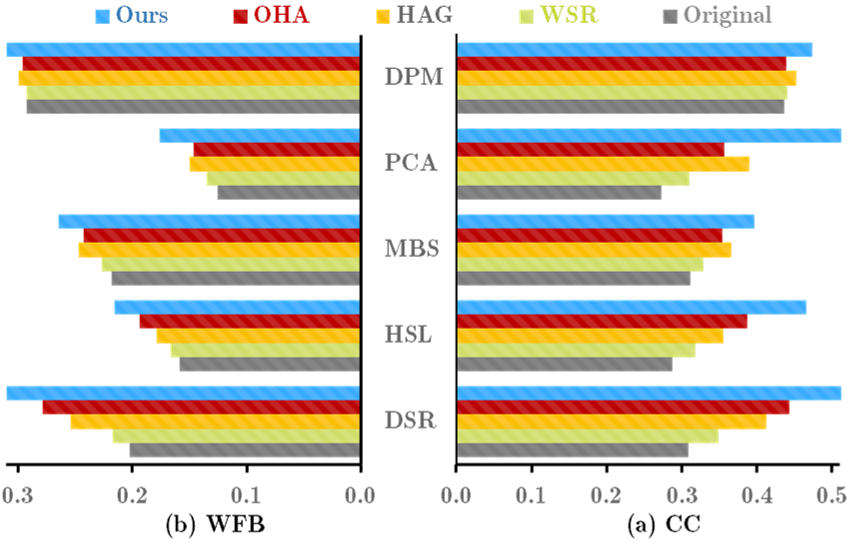}
    \vspace{-0.2cm}
    \caption{\textbf{Enhancement evaluation:} The bars represent the (right) Correlation-Coefficient (CC) and (left) the Weighted F-beta (WFB)~\protect\cite{margolin2014evaluate} scores obtained when comparing the ground-truth masks with saliency maps computed using five different saliency estimation algorithms (see text). The longer the bar, the more similar the saliency maps are to the ground-truth. It can be seen that the saliency maps of our manipulated images are consistently more similar to the ground-truth.}
  \vspace{-0.2cm}
  \label{fig:evaluate}
\end{figure}

\begin{figure}[t]
	\centering
	\includegraphics[width=.9\linewidth]{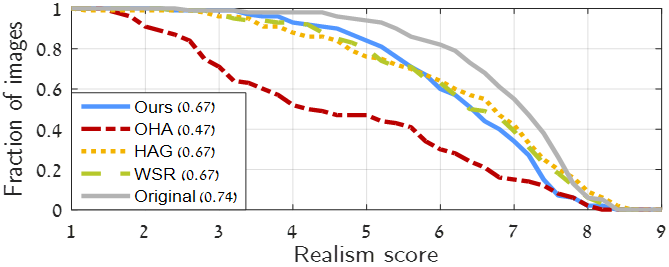}
    \vspace{-0.1cm}
\caption{\textbf{Realism evaluation.} Realism scores obtained via a user survey (see text for details). The curves show the fraction of images with average score greater than \emph{Realism score}. The Area-Under-Curve (AUC) values are presented in the legend.
Our manipulated images are ranked as more realistic than those of OHA and similar to those of WSR and HAG. this is while our enhancement effects are more robust, as shown in Figure~\ref{fig:enhance}. }
\vspace{-0.2cm}
  \label{fig:realism}
\end{figure}

\begin{figure*}[ht!]
		\centering
		\rotatebox{90}{\hspace{0.1cm}}~
		\vspace{0.1cm}
		\begin{subfigure}[b]{.23\textwidth}\includegraphics[width=.98\linewidth]{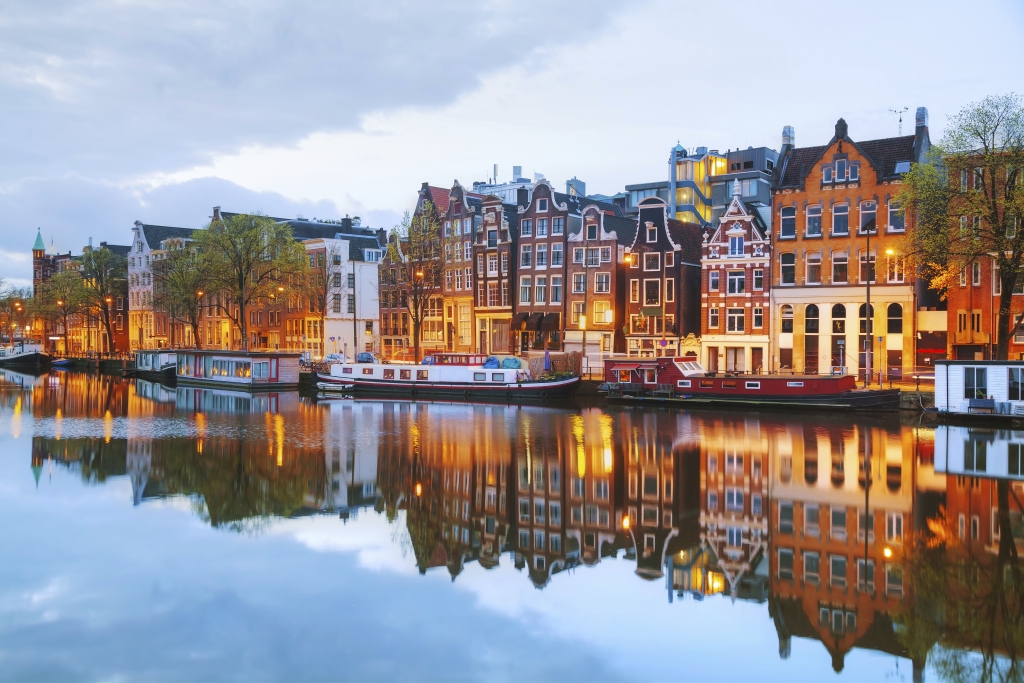}\end{subfigure}~
		\begin{subfigure}[b]{.23\textwidth}\includegraphics[width=.98\linewidth]{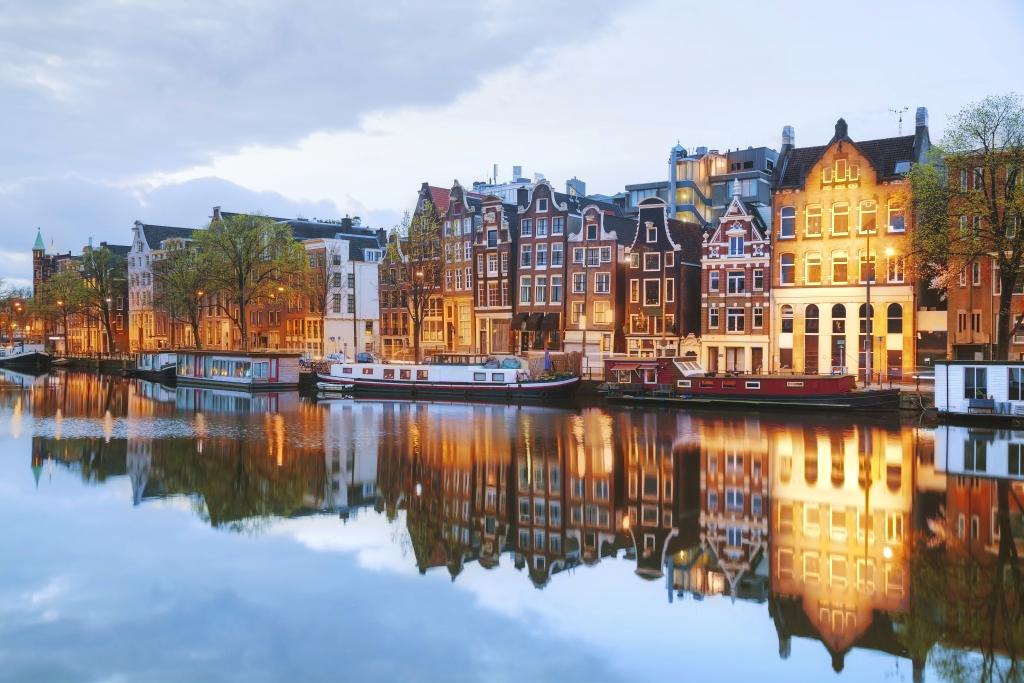}\end{subfigure}~
		\begin{subfigure}[b]{.23\textwidth}\includegraphics[width=.98\linewidth]{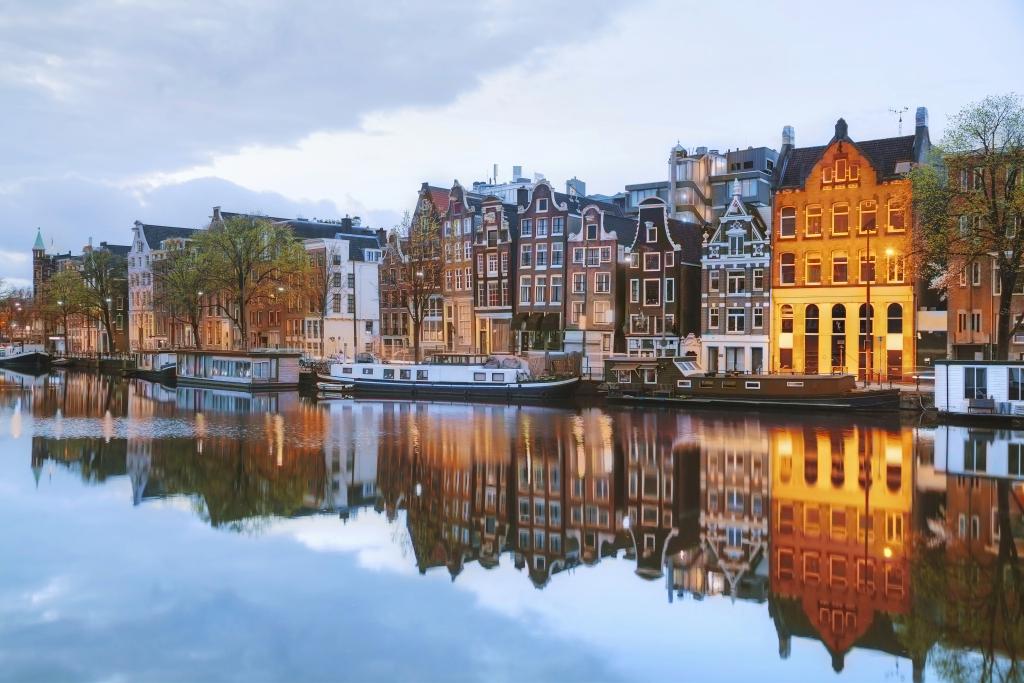}\end{subfigure}~
		\begin{subfigure}[b]{.23\textwidth}\includegraphics[width=.98\linewidth]{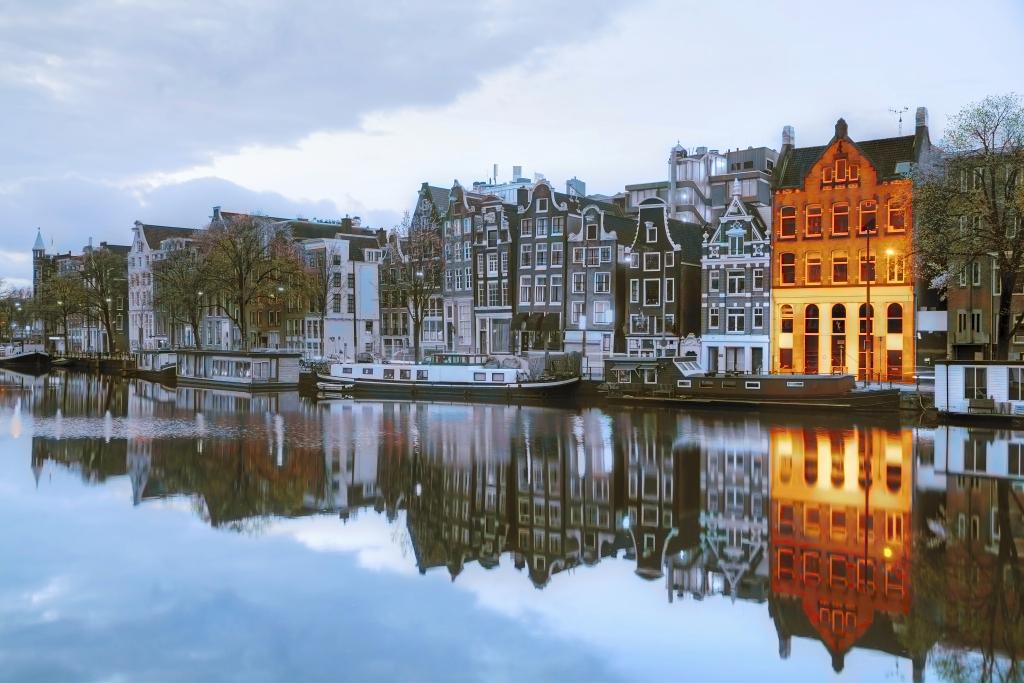}\end{subfigure}
		\rotatebox{90}{\hspace{0.1cm}}~
		\vspace{-0.2cm}
		\begin{subfigure}[b]{.23\textwidth}\includegraphics[width=.98\linewidth]{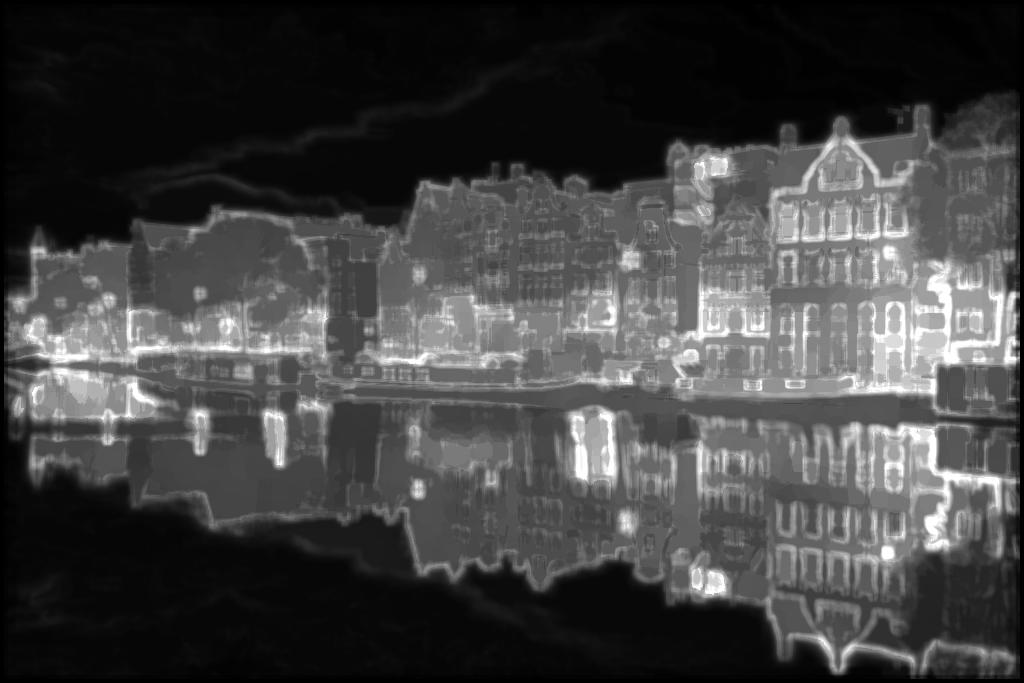}\caption{Input Image}\end{subfigure}~
		\begin{subfigure}[b]{.23\textwidth}\includegraphics[width=.98\linewidth]{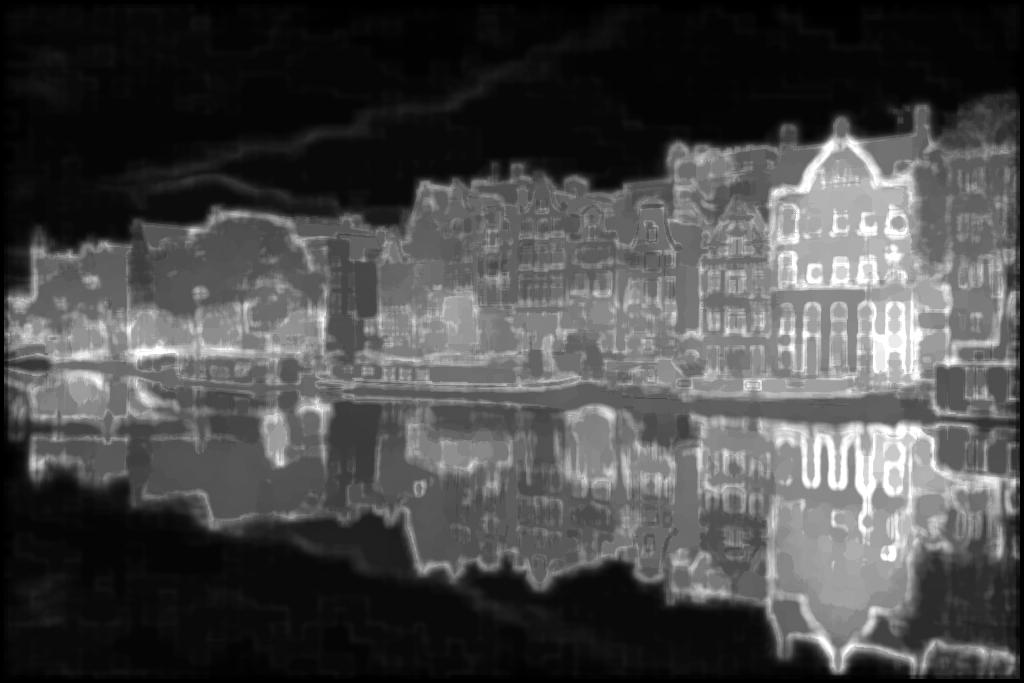}\caption{$\Delta S=0.4$}\end{subfigure}~
		\begin{subfigure}[b]{.23\textwidth}\includegraphics[width=.98\linewidth]{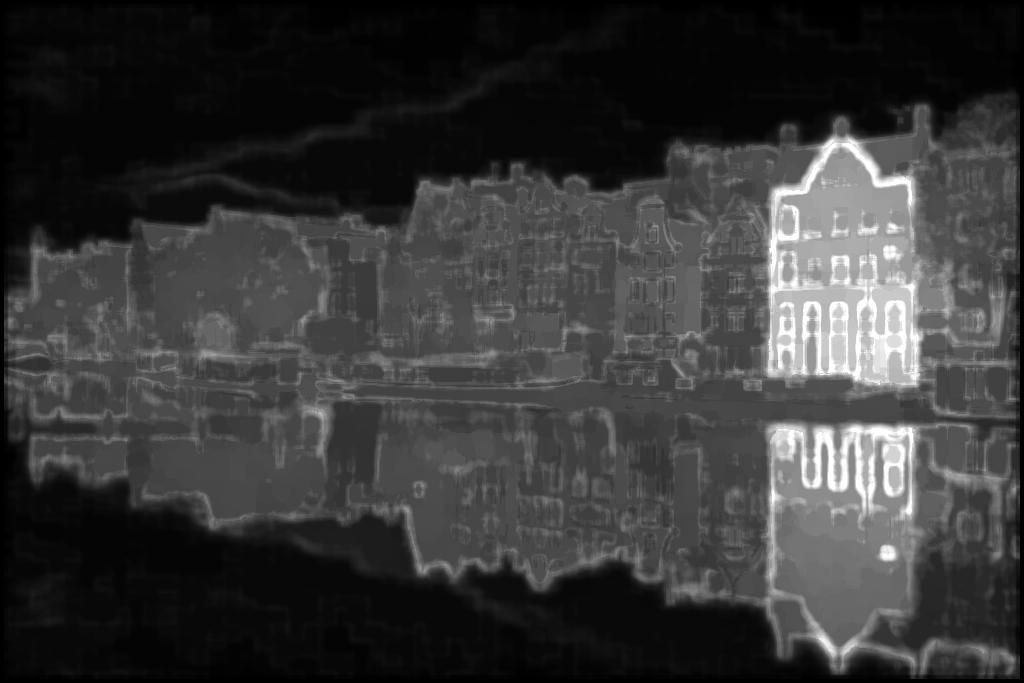}\caption{$\Delta S=0.6$}\end{subfigure}~
		\begin{subfigure}[b]{.23\textwidth}\includegraphics[width=.98\linewidth]{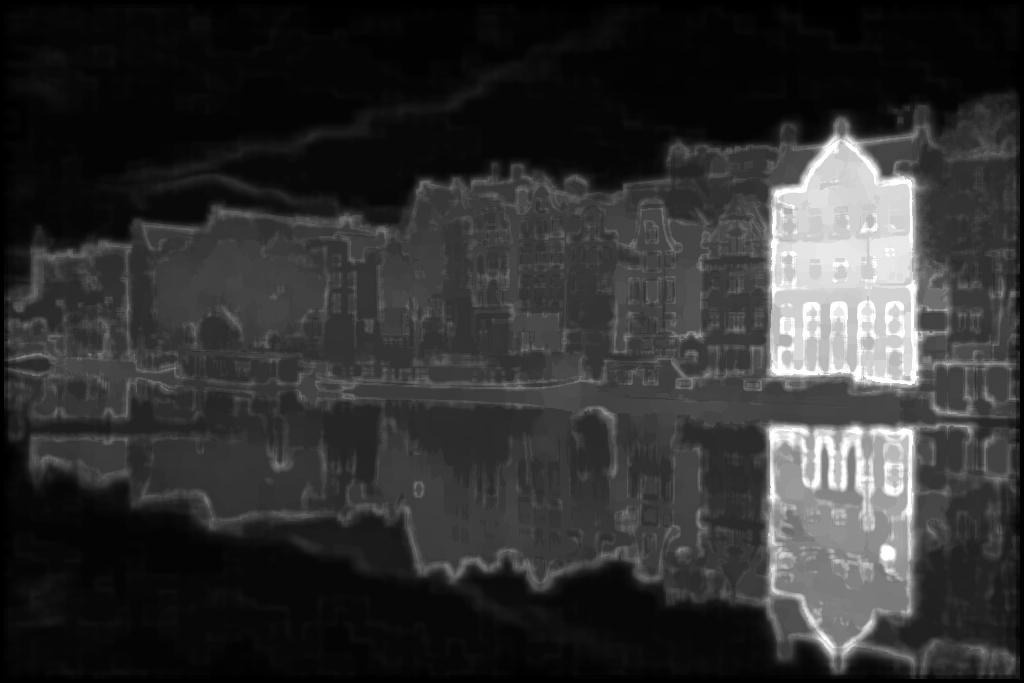}\caption{$\Delta S=0.8$}\end{subfigure}
		\caption{\textbf{Controlling the level of enhancement.} (Top) (a) Input image. (b,c,d) The manipulated image $J$ with $\Delta S=0.4,0.6,0.8$, respectively. (Bottom) the corresponding saliency maps. As $\Delta S$ is increased, so does the saliency contrast between the foreground and the background. As mask, the user marked the rightmost house and its reflection on the water. }
        \vspace{-0.2cm}
		\label{fig:deltas}
\end{figure*}

\textbf{Enhancement evaluation:}
To measure how successful a manipulated image is, we do the following. We take the user provided mask as the ground-truth saliency map. We then compute the saliency map of the manipulated image and compare it to the ground-truth.
To provide a reliable assessment we use five different salient object detection methods: MBS~\cite{zhang2015minimum}, HSL~\cite{yan2013hierarchical}, DSR~\cite{li2013saliency}, PCA~\cite{margolin2013makes} and MDP\cite{LiYu15}, each based on different principles (patch based, CNN, geodesic distance etc.).
The computed saliency maps are compared to the ground-truth using two commonly-used metrics for saliency evaluation: (i) Pearson’s-Correlation-Coefficient (CC) which was recommended by~\cite{bylinskii2016different} as the best option for assessing saliency maps, and, (ii) Weighted F-beta (WFB)~\cite{margolin2014evaluate} which was shown to be a preferred choice for evaluation of foreground maps. 

The bar plots in Figure~\ref{fig:evaluate} show that the saliency maps of our manipulated images are more similar to the ground-truth than those of OHA, WSR and HAG. This is true for both saliency measures and for all five methods for saliency estimation.

\textbf{Realism:} As mentioned earlier, being able to enhance a region does not suffice. We must also verify that the manipulated images look plausible and realistic. We measure this via a user survey. Each image was presented to human participants who were asked a simple question: ``Does the image look realistic?'' The scores were given on a scale of [1-9], where 9 is 'definitely realistic' and 1 is 'definitely unrealistic'. We used Amazon Mechanical Turk to collect 20 annotations per image, where each worker viewed only one version of each image out of five. 
The survey was performed on a random subset of 20\% of the data-set.

Figure~\ref{fig:realism} shows for each enhancement method the fraction of images with average score larger than a realism score $\in [1,9]$ and the overall AUC values. 
OHA results are often non-realistic, which is not surprising given their approach uses colors far from those in the original image.
Our manipulated images are mostly realistic and similar to WSR and HAG in the level of realism. Recall, that this is achieved while our success in obtaining the desired enhancement effect is much better.

\textbf{Controlling the Level of Enhancement:} One of the advantages of our approach over previous ones is the control we provide the user over the degree of the manipulation effect.
Our algorithm accepts a single parameter from the user, $\Delta S$, which determines the level of enhancement. The higher $\Delta S$ is, the more salient will the region of interest become, since our algorithm  minimizes $E^{sal}$, i.e., it aims to achieve $\psi (S_J,R) = \Delta S$. While we chose $\Delta S=0.6$ for most images, another user could prefer other values to get more or less prominent effects. Figure~\ref{fig:deltas} illustrates the influence $\Delta S$ on the manipulation results.

\textit{The user-provided mask:}
In our dataset, the mask was marked by users to define a salient object in the scene. In order to use our method on a new image the user is required to mark the region that input region. Note that similarly to other imaging tasks, such as, image completion, compositing, recoloring and warping, the definition of the target region is up to the user to determine and is not part of the method. To facilitate the selection the user can utilize interactive methods such as \cite{long2015fully,rother2004grabcut,xu2016deep} to easily generate region-of-interest masks.

\subsection{Other Applications}
\begin{figure}[htb]
		\centering
		\rotatebox{90}{\hspace{0.1cm}}~
		\vspace{0cm}
		\begin{subfigure}[b]{.15\textwidth}\includegraphics[width=.98\linewidth]{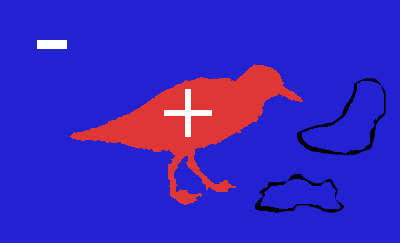}
        \caption{}
        \label{fig:setupsa}\end{subfigure}~
		\begin{subfigure}[b]{.15\textwidth}\includegraphics[width=.98\linewidth]{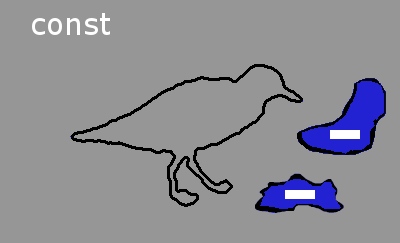}
        \caption{}
        \end{subfigure}~
		\begin{subfigure}[b]{.15\textwidth}\includegraphics[width=.98\linewidth]{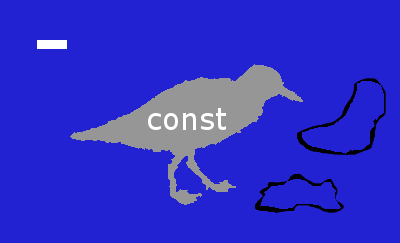}
        \caption{}
        \end{subfigure}
		\vspace{-0.7cm}
		\caption{\textbf{Mask setups.} Illustration of the setups used for: (a) object enhancement, (b)  distractor attenuation and (c) decluttering. We increase the saliency in \textcolor[rgb]{1,0,0}{red}, decrease it in \textcolor[rgb]{0,0,1}{blue} and apply no change in \textcolor[rgb]{0.3,0.3,0.3}{gray}.}
		\label{fig:setups}
\end{figure}

\vspace*{-0.1cm}
Since our framework allows both increasing and decreasing of saliency it enables two additional applications: 
(i) \emph{Distractor Attenuation}, where the target's saliency is decreased, and (ii) \emph{Background Decluttering}, where the target is unchanged while salient pixels in the background are demoted. 
A nice property of our approach is that all that is required for these is using a different mask setup, as illustrated in Figure~\ref{fig:setups}.

\begin{figure*}[tb!]
		\centering
		\rotatebox{90}{\hspace{0.1cm}}~
		\vspace{0.1cm}
		\begin{subfigure}[t]{0.2\textwidth}\includegraphics[height=2.5cm]{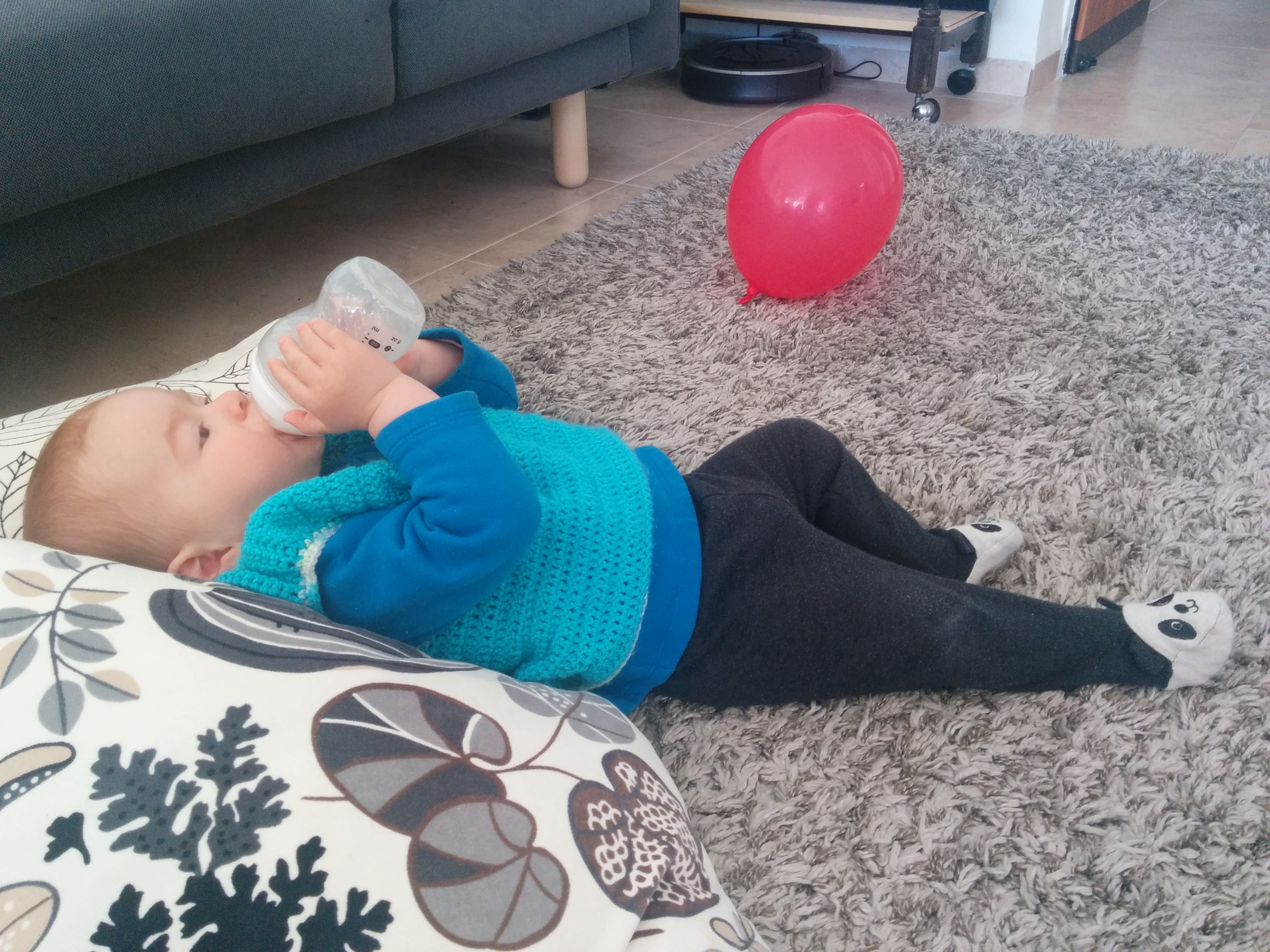}\end{subfigure}~
		\begin{subfigure}[t]{0.2\textwidth}\includegraphics[height=2.5cm]{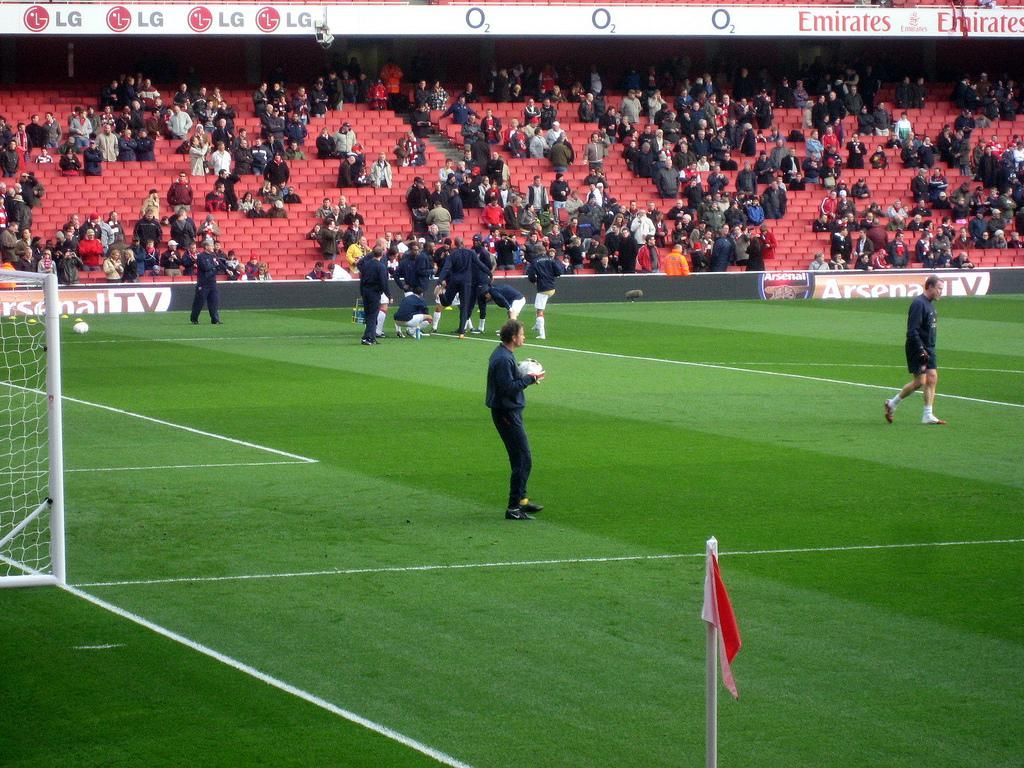}\end{subfigure}~
		\begin{subfigure}[t]{0.19\textwidth}\includegraphics[height=2.5cm]{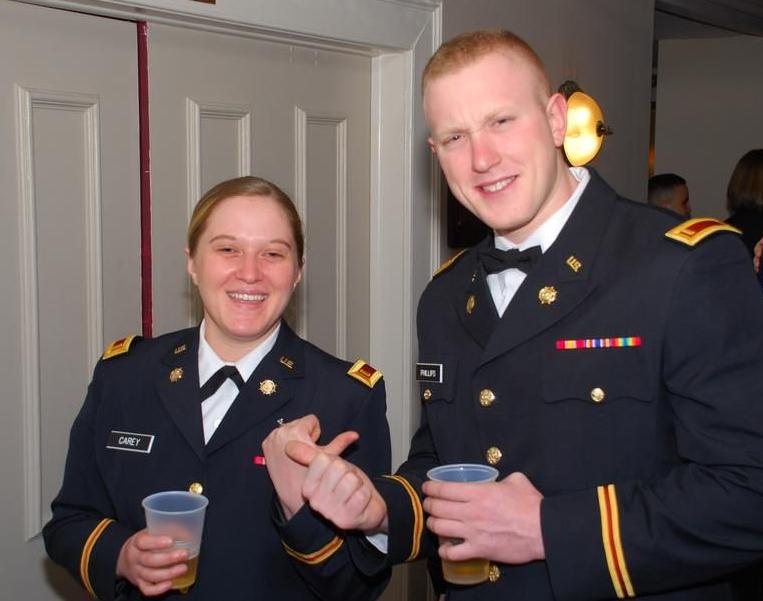}\end{subfigure}~
		\begin{subfigure}[t]{0.12\textwidth}\includegraphics[height=2.5cm]{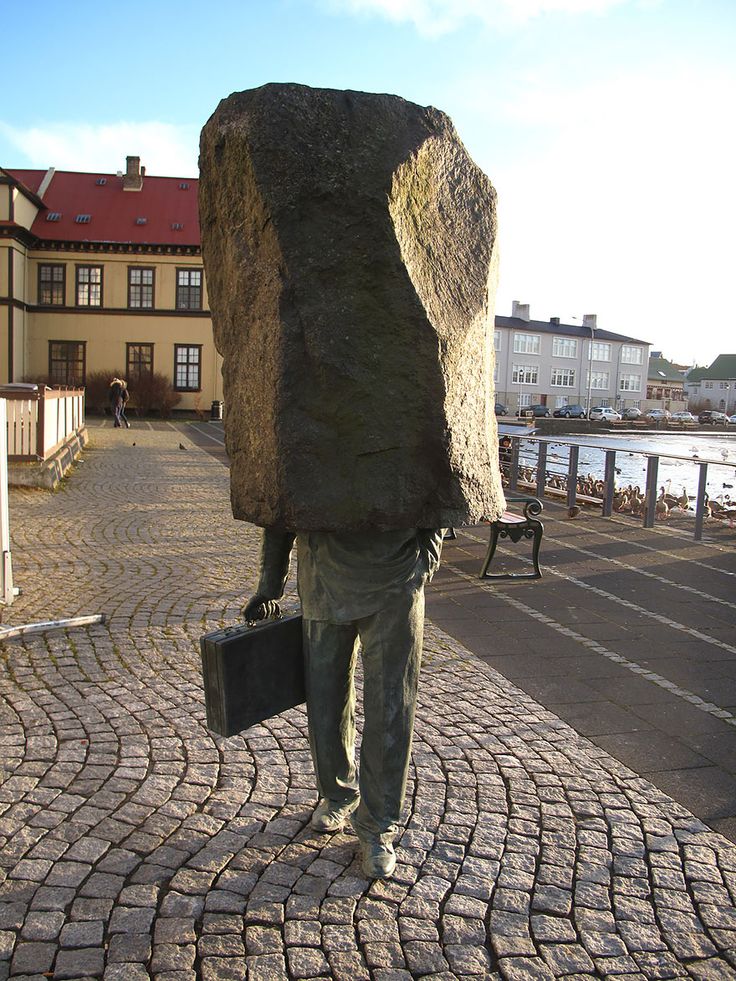}\end{subfigure}~
		\begin{subfigure}[t]{0.12\textwidth}{\includegraphics[height=2cm]{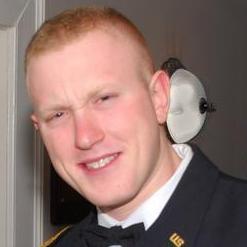}}\end{subfigure}~
    \begin{subfigure}[t]{0.12\textwidth}{\includegraphics[height=2cm]{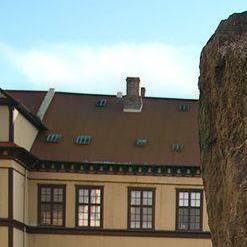}}\end{subfigure}
		\rotatebox{90}{\hspace{0.1cm}}~
		\vspace{-0.2cm}
		\begin{subfigure}[t]{0.2\textwidth}\includegraphics[height=2.5cm]{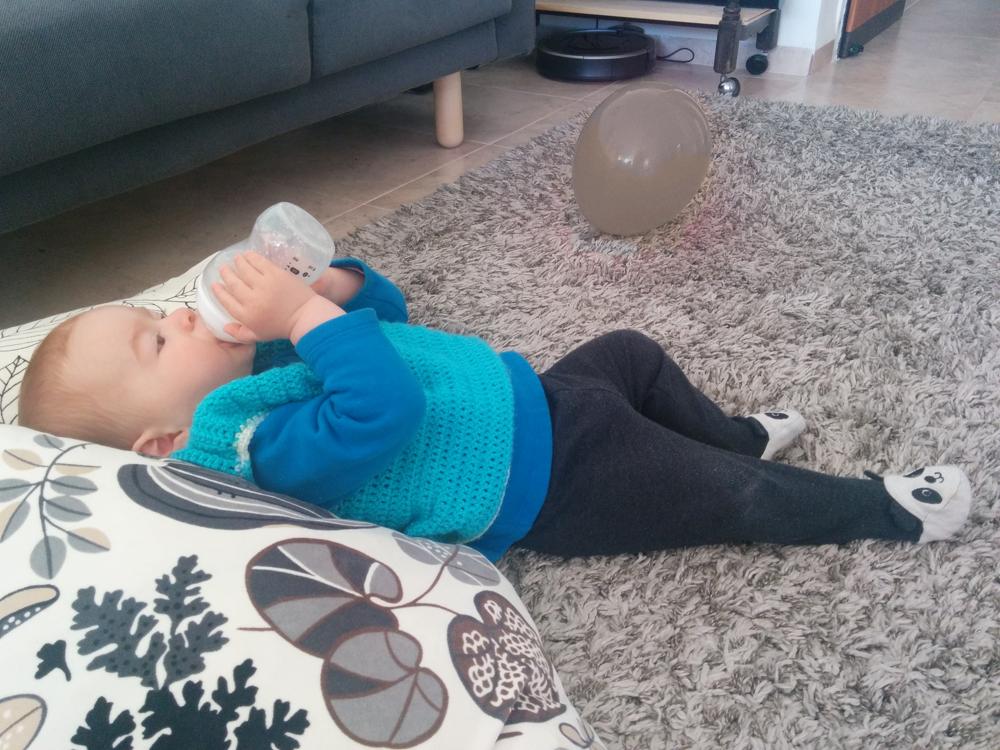}\caption{Example 1}\end{subfigure}~
		\begin{subfigure}[t]{0.2\textwidth}\includegraphics[height=2.5cm]{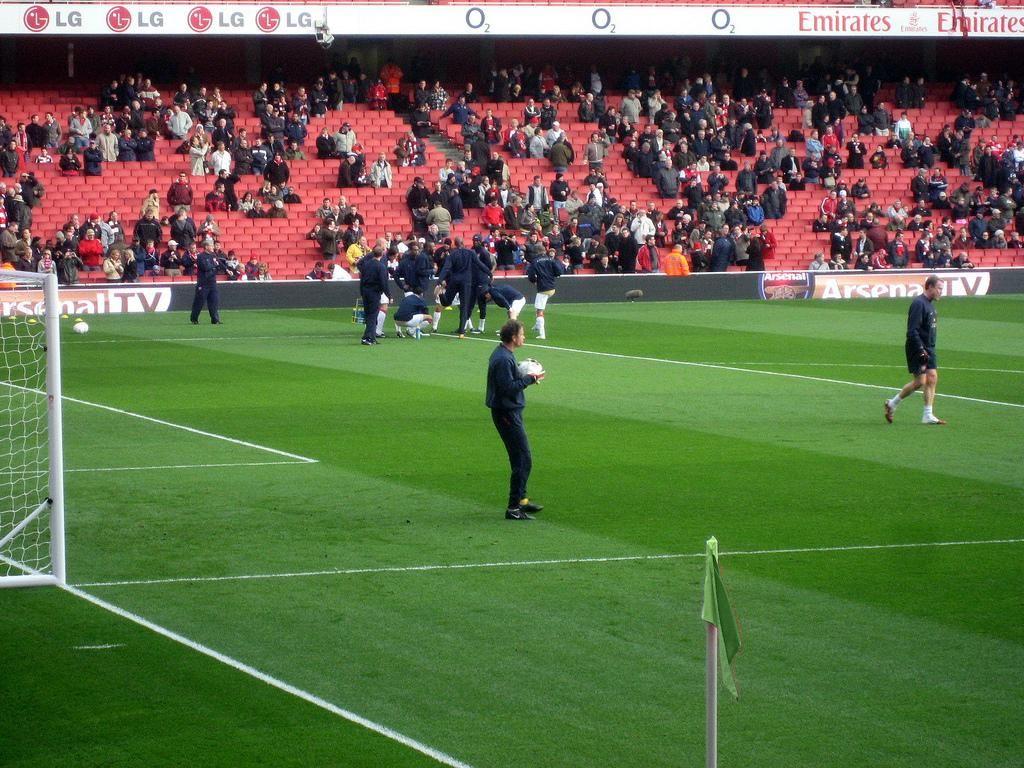}\caption{Example 2}\end{subfigure}~
		\begin{subfigure}[t]{0.19\textwidth}\includegraphics[height=2.5cm]{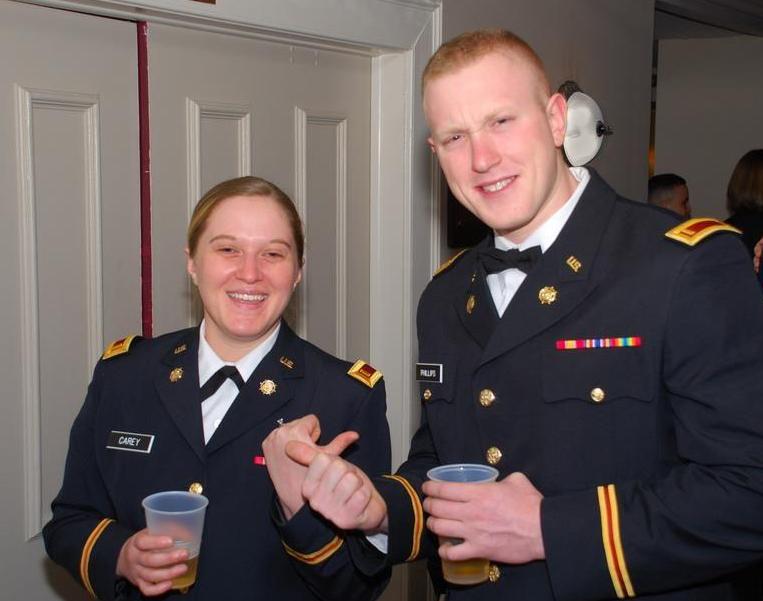}\caption{Example 3}\end{subfigure}~
    \begin{subfigure}[t]{0.12\textwidth}\includegraphics[height=2.5cm]{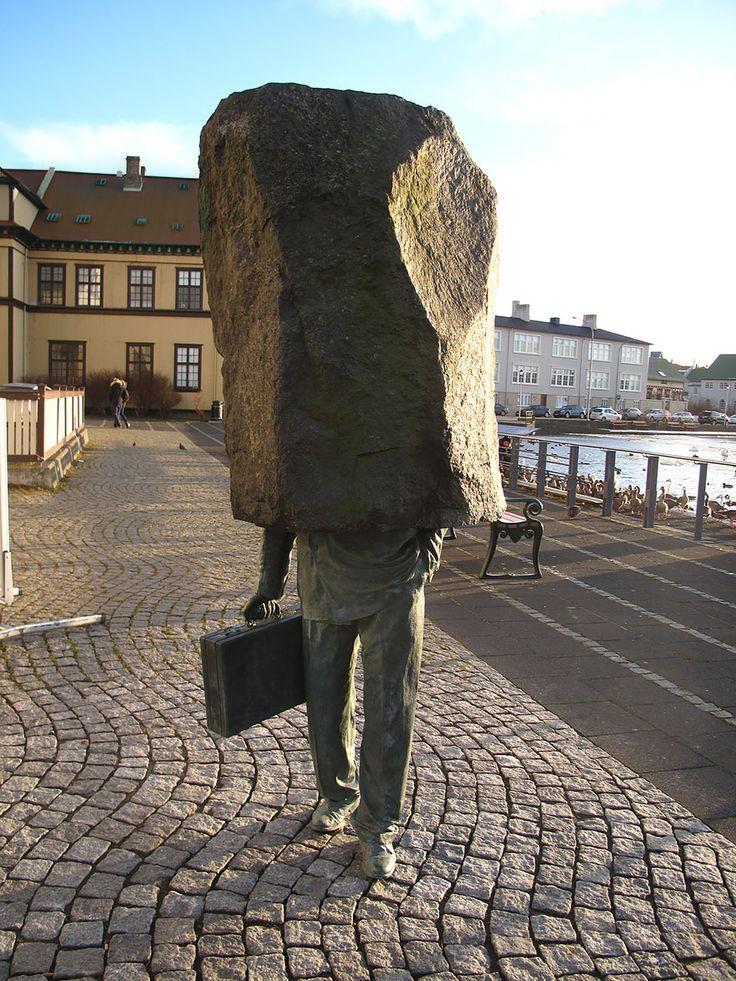}\caption{Example 4}\end{subfigure}~
		\begin{subfigure}[t]{0.12\textwidth}{\includegraphics[height=2cm]{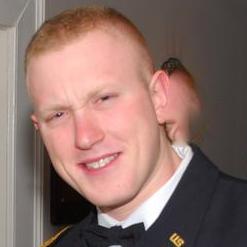}}\caption{Inpainting 1}\end{subfigure}~
		\begin{subfigure}[t]{0.12\textwidth}{\includegraphics[height=2cm]{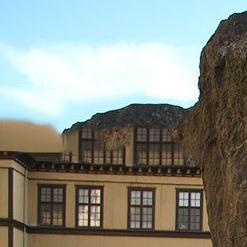}}\caption{Inpainting 2}\end{subfigure}
		\caption{\textbf{Distractor Attenuation.} (a)-(d) Top: Input images. The distractors were the balloon, the red flag, the shiny lamp and the red roof. Bottom: our manipulated images after reducing the saliency of the distractors. (e)-(f) Top: Zoom in on our result. Bottom: Zoom in on the inpainting result by Adobe Photoshop showing typical artifacts of inpainting methods.}
        \vspace{0.2cm}
		\label{fig:distractors}
        \vspace*{-0.3cm}
\end{figure*}		
\vspace*{-0.2cm}

\begin{figure*}
		\centering
		\rotatebox{90}{\hspace{0.1cm}}~
		\vspace{-0.2cm}
		\begin{subfigure}[b]{.21\textwidth}\includegraphics[width=.98\linewidth]{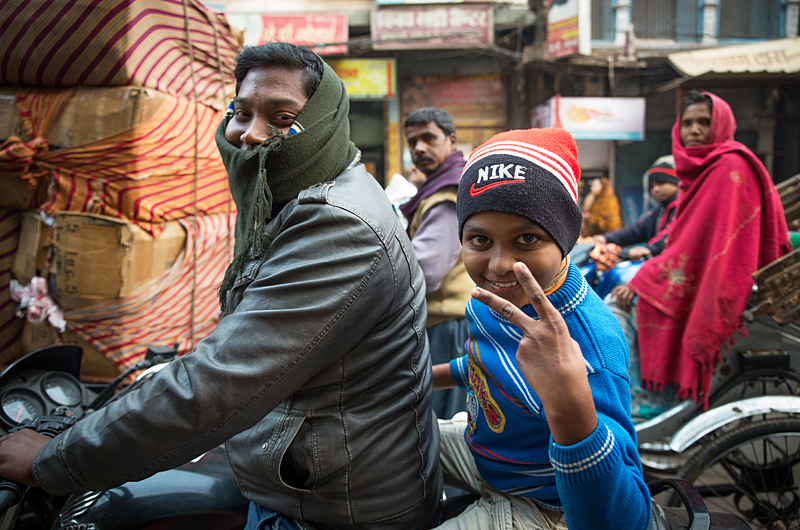}\caption{Input image}\end{subfigure}~
		\begin{subfigure}[b]{.21\textwidth}\includegraphics[width=.98\linewidth]{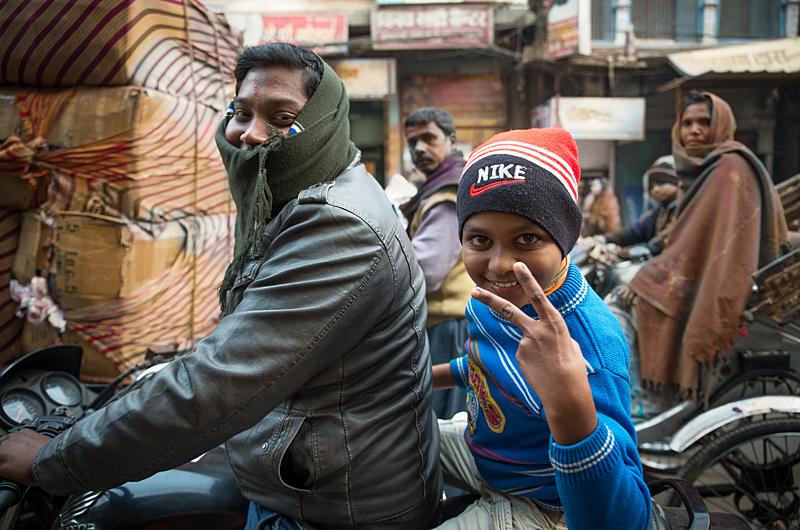}\caption{Manipulated image}\end{subfigure}~
		\begin{subfigure}[b]{.21\textwidth}\includegraphics[width=.98\linewidth]{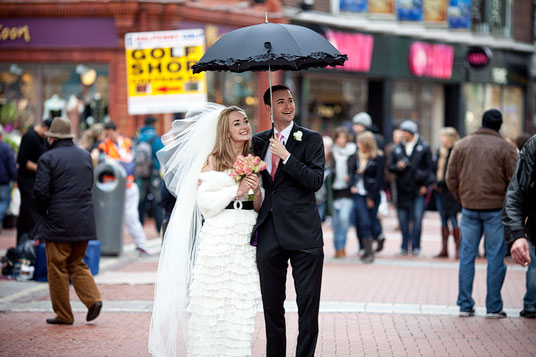}\caption{Input image}\end{subfigure}~
		\begin{subfigure}[b]{.21\textwidth}\includegraphics[width=.98\linewidth]{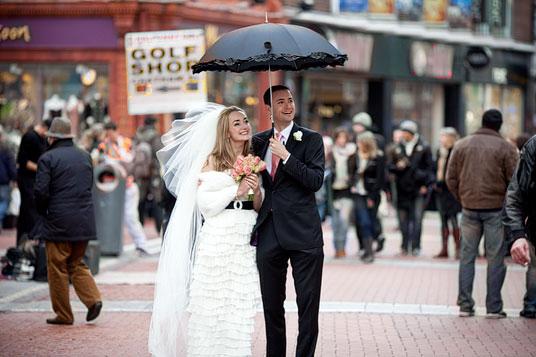}\caption{Manipulated image}\end{subfigure}
		\begin{subfigure}[b]{.115\textwidth}\includegraphics[width=.74\linewidth]{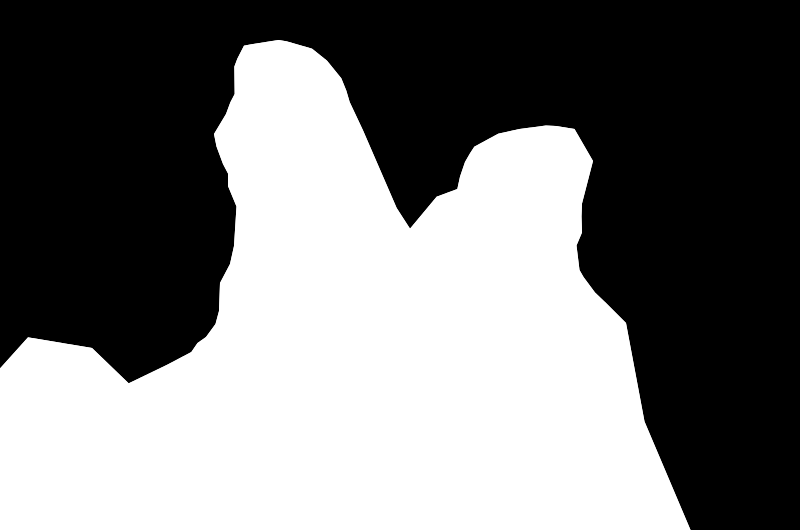}\vspace{0.37cm}\\ 		
\includegraphics[width=.74\linewidth]{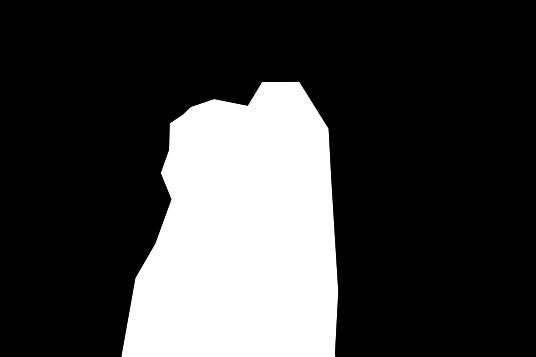}\caption{Masks}\end{subfigure}
		\caption{\textbf{Background DeCluttering.} Often in cluttered scenes one would like to reduce the saliency of background regions to get a less noisy image. In such cases it suffices to loosely mark the foreground region as shown in (e), since the entire background is manipulated. In (a,b) saliency was reduced for the boxes on the left and red sari on the right. In (c,d) the signs in the background were demoted thus drawing attention to the bride and groom.}
		\label{fig:denoising}
\vspace*{-0.3cm}
\end{figure*}

\textbf{Distractor Attenuation:} The task of getting rid of distractors was recently defined by Fried et al.~\cite{fried2015finding}. Distractors are small localized regions that turned out salient against the photographer's intentions. In~\cite{fried2015finding} distractors were removed entirely from the image and the holes were filled by inpainting. This approach has two main limitations. First, it completely removes objects from the image thus changing the scene in an obtrusive manner that might not be desired by the user. Second, hole-filling methods hallucinate data and sometimes produce weird effects.

Instead, we propose to keep the distractors in the image while reducing their saliency. Figure~\ref{fig:distractors} presents some of our results and comparisons to those obtained by inpainting. We succeed to attenuate the saliency of the distractors, without having to remove them from the image.

\textbf{Background Decluttering:}
Reducing saliency is also useful for images of cluttered scenes where one's gaze dynamically shifts across the image to spurious salient locations in the background. Some examples of this phenomena and how we attenuate it are presented in Figure~\ref{fig:denoising}. 
This scenario resembles that of removing distractors, with one main difference. Distractors are usually small localized objects, therefore, one could potentially use inpainting to remove them. Differently, when the background is cluttered, marking all the distractors could be tedious and removing them would result in a completely different image.

Our approach easily deals with cluttered background. The user is requested to loosely mark the foreground region. We then leave the foreground unchanged and manipulate only the background, using $\DB^{-}$ to automatically decrease the saliency of clutter pixels. The optimization modifies only background pixels with high saliency, since those with low saliency are represented in $\DB^{-}$ and therefore are matched to themselves.

\begin{figure*}[h]
		\centering
    \rotatebox{90}{masks\hspace{0.3cm}}~
	\vspace{0.1cm}
    \begin{subfigure}[b]{.1225\textwidth}\includegraphics[width=\textwidth]{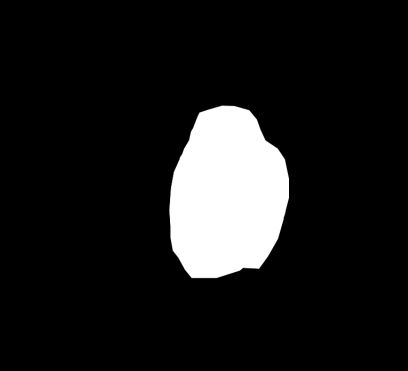}\end{subfigure}~
    \begin{subfigure}[b]{.1225\textwidth}\includegraphics[width=\textwidth]{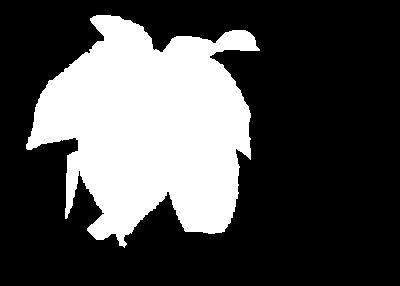}\end{subfigure}~
	\begin{subfigure}[b]{.1225\textwidth}\includegraphics[width=\textwidth]{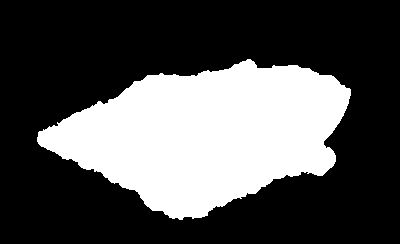}\end{subfigure}~
    \begin{subfigure}[b]{.1225\textwidth}\includegraphics[width=\textwidth]{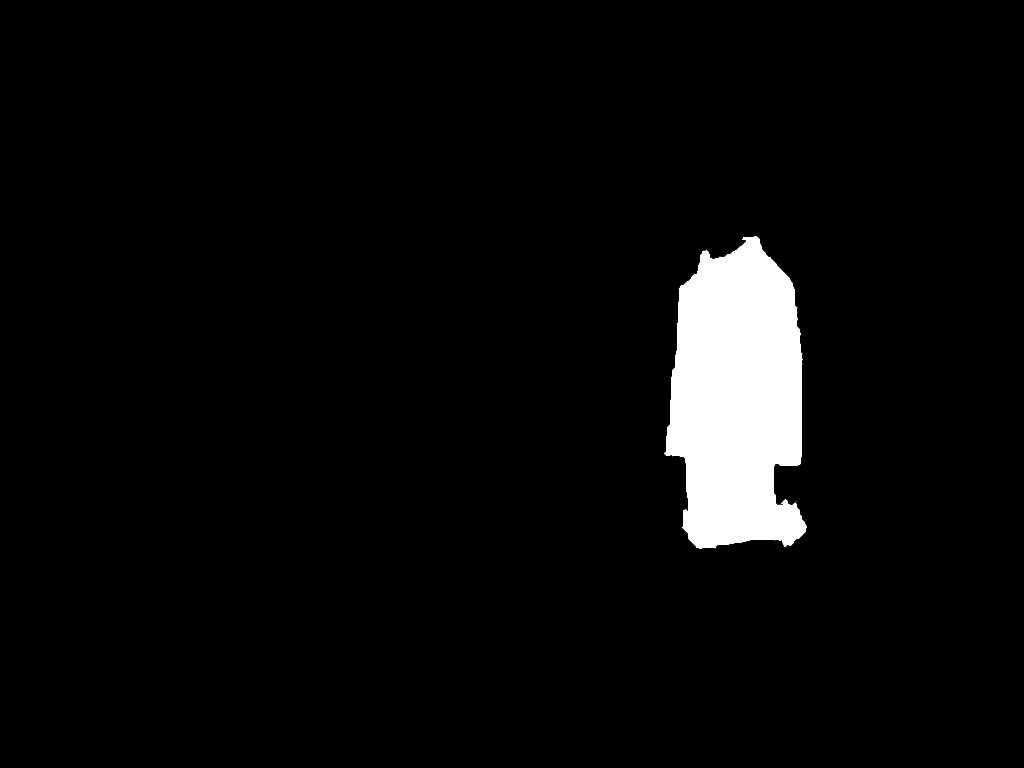}\end{subfigure}~
    \begin{subfigure}[b]{.1225\textwidth}\includegraphics[width=\textwidth]{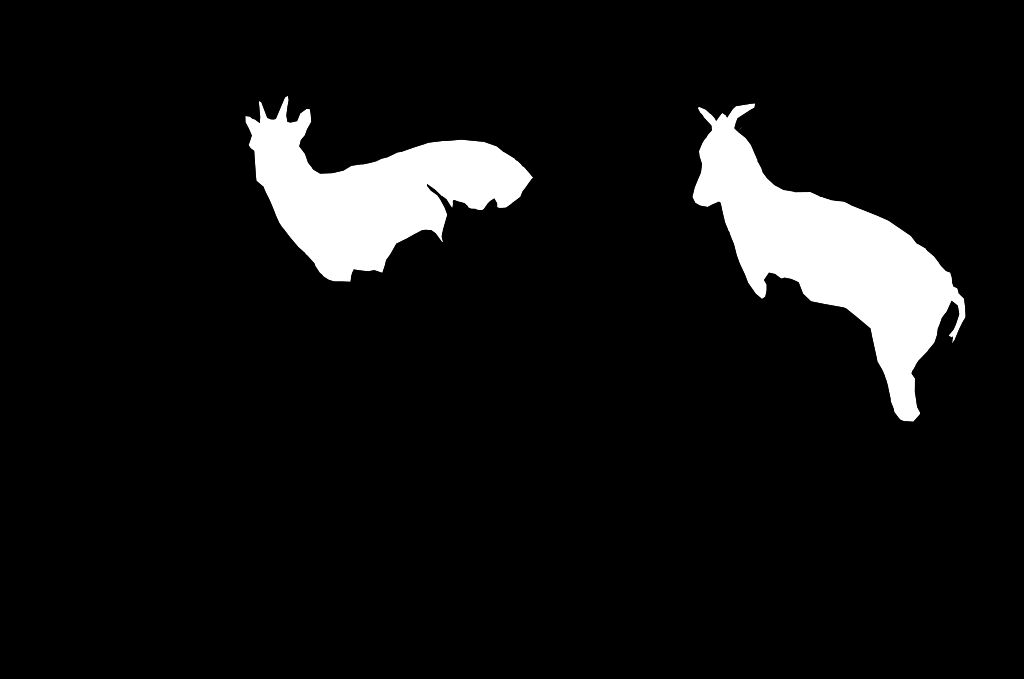}\end{subfigure}~
    \begin{subfigure}[b]{.1225\textwidth}\includegraphics[width=\textwidth]{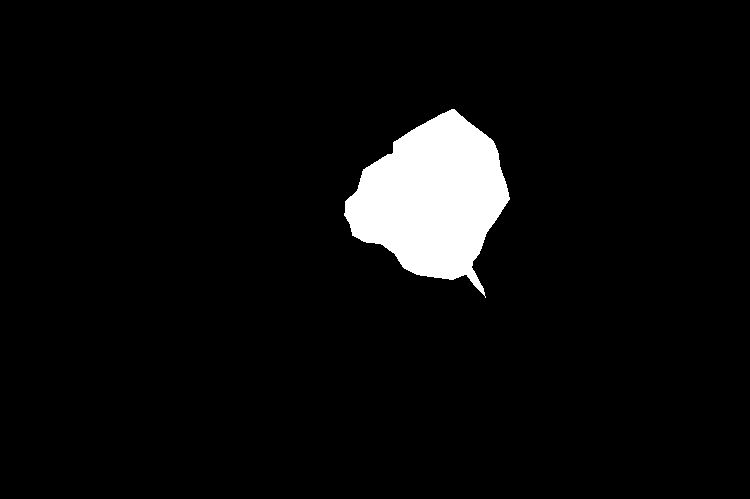}\end{subfigure}~
    \begin{subfigure}[b]{.1225\textwidth}\includegraphics[width=\textwidth]{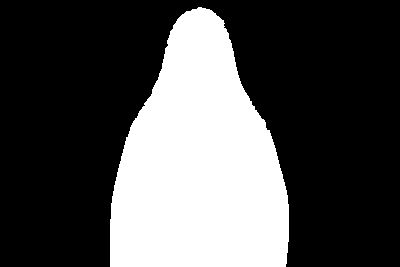}\end{subfigure}
	\rotatebox{90}{\hspace{0.1cm}}~
	\vspace{0.1cm}\\
    \begin{tabular}{cccccc}    
    (1)&
  \includegraphics[width=.18\textwidth]{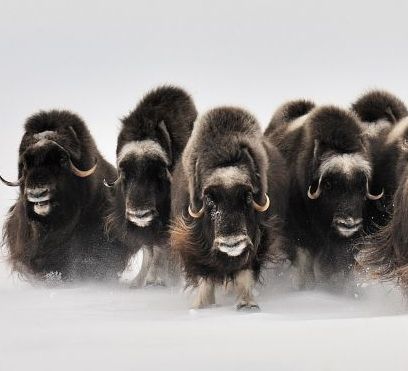}&
  \includegraphics[width=.18\textwidth]{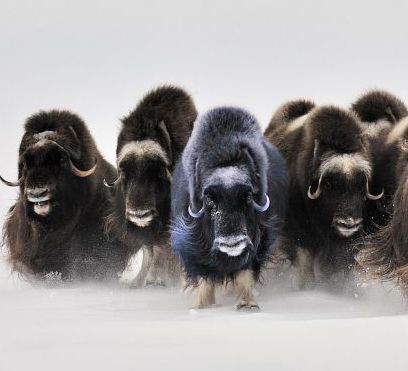}&
  \includegraphics[width=.18\textwidth]{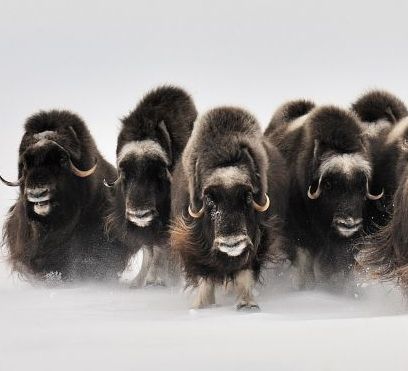}&
  \includegraphics[width=.18\textwidth]{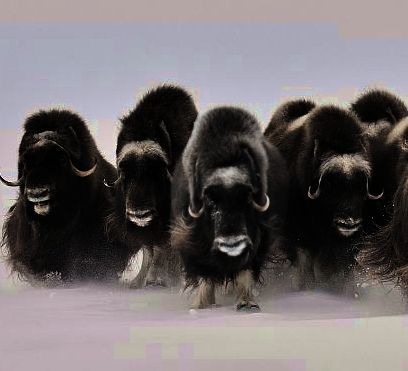}&
  \includegraphics[width=.18\textwidth]{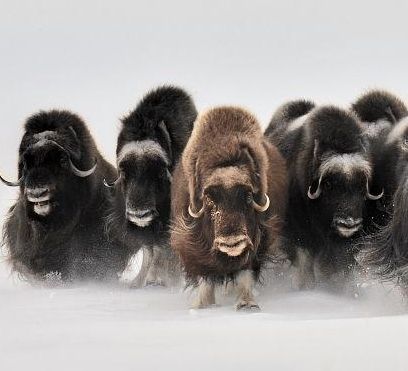}\\
        (2)&
        \includegraphics[width=.18\textwidth]{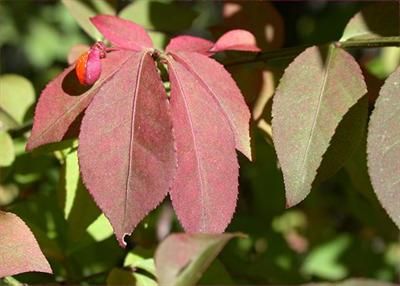}&
        \includegraphics[width=.18\textwidth]{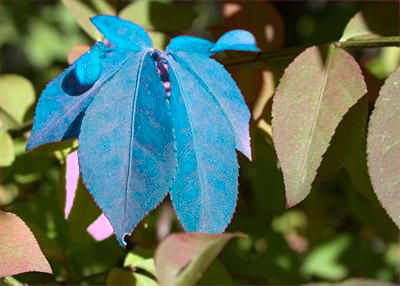}&
        \includegraphics[width=.18\textwidth]{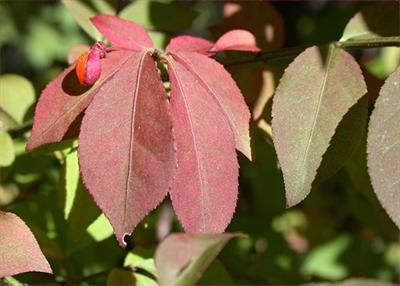}&
        \includegraphics[width=.18\textwidth]{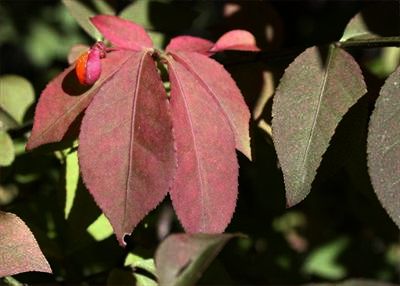}&
        \includegraphics[width=.18\textwidth]{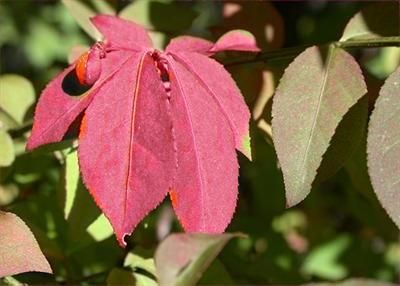}\\
        (3)&
        \includegraphics[width=.18\textwidth]{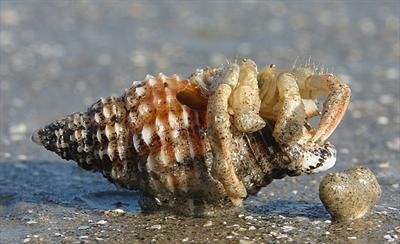}&
        \includegraphics[width=.18\textwidth]{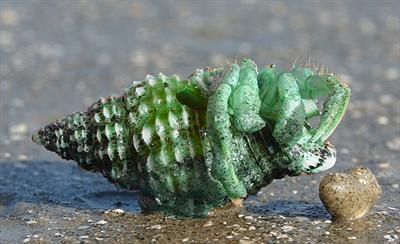}&
        \includegraphics[width=.18\textwidth]{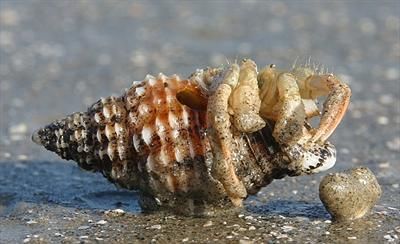}&
        \includegraphics[width=.18\textwidth]{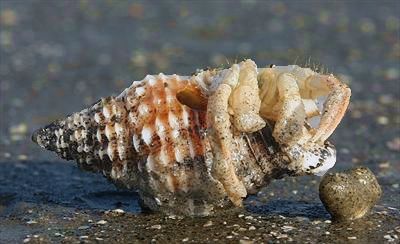}&
        \includegraphics[width=.18\textwidth]{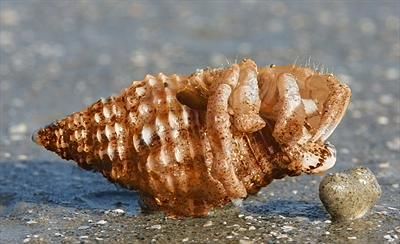}\\
        (4)&
 \includegraphics[width=.18\textwidth]{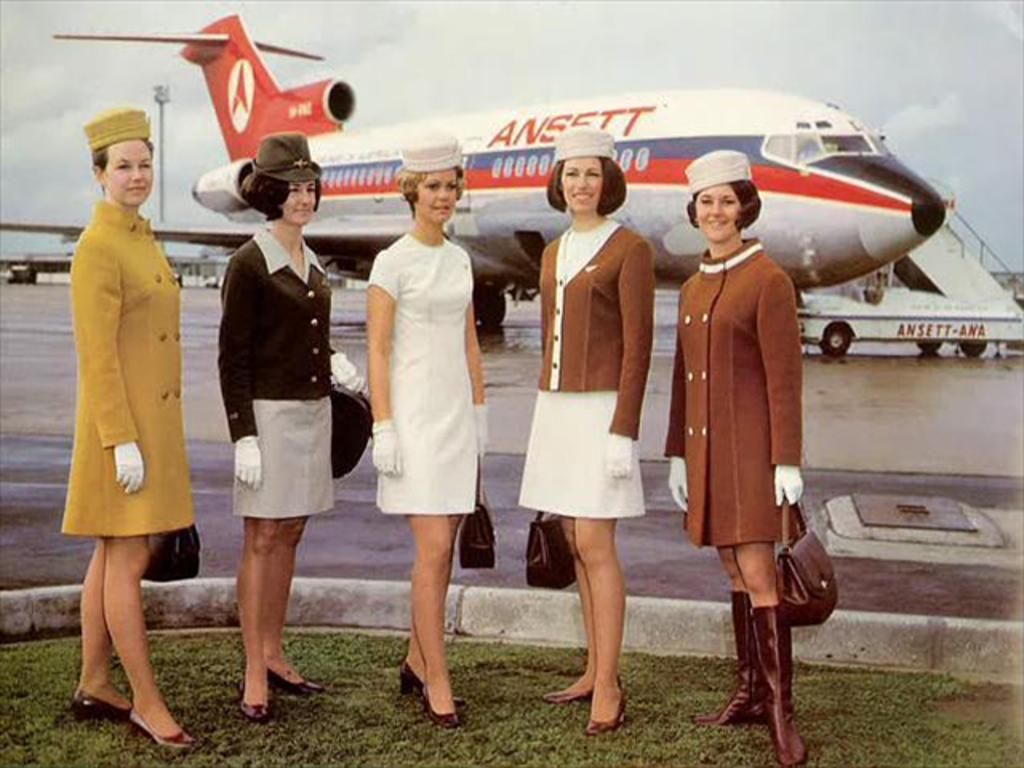}&
 \includegraphics[width=.18\textwidth]{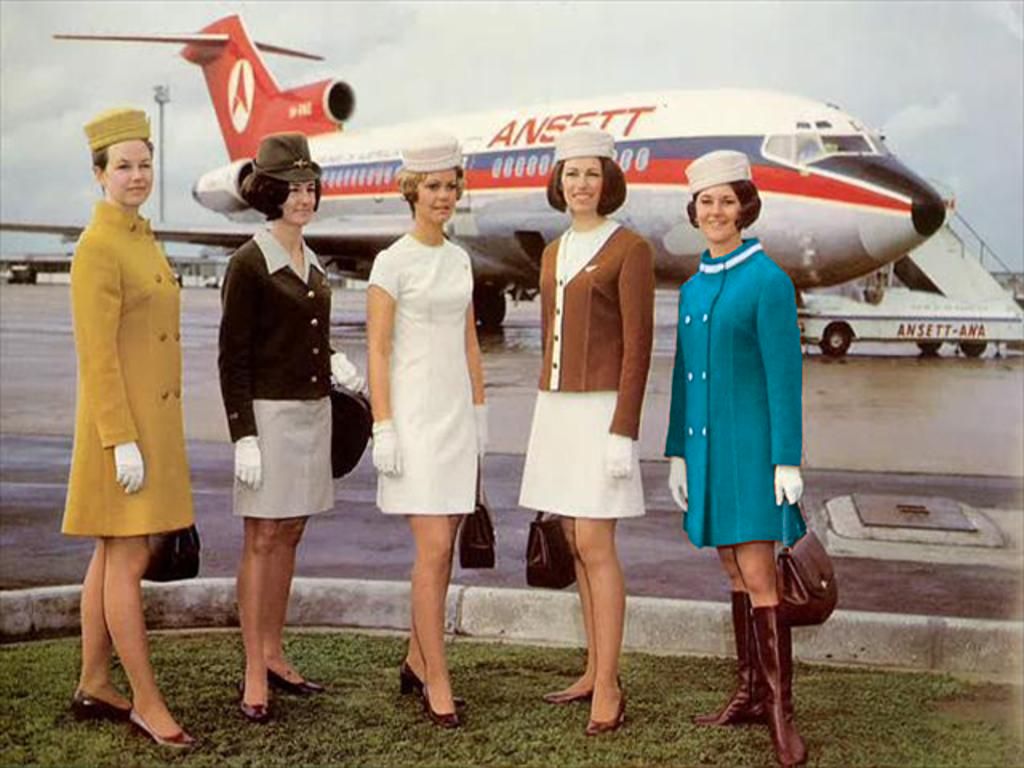}&
 \includegraphics[width=.18\textwidth]{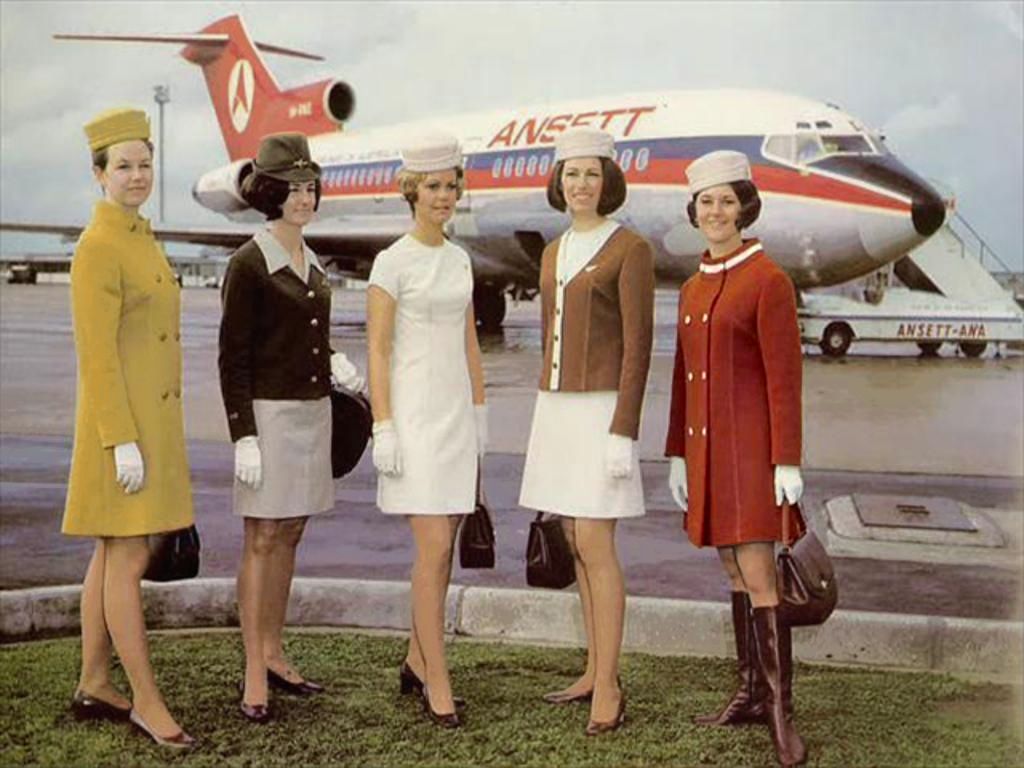}&
        \includegraphics[width=.18\textwidth]{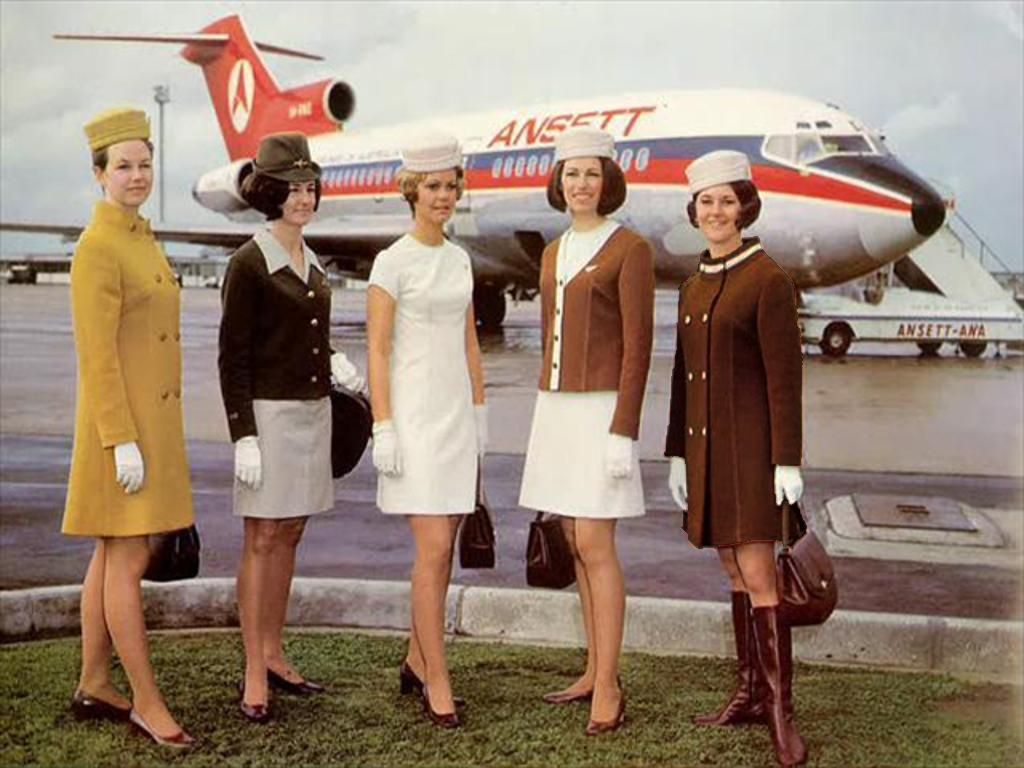}&
        \includegraphics[width=.18\textwidth]{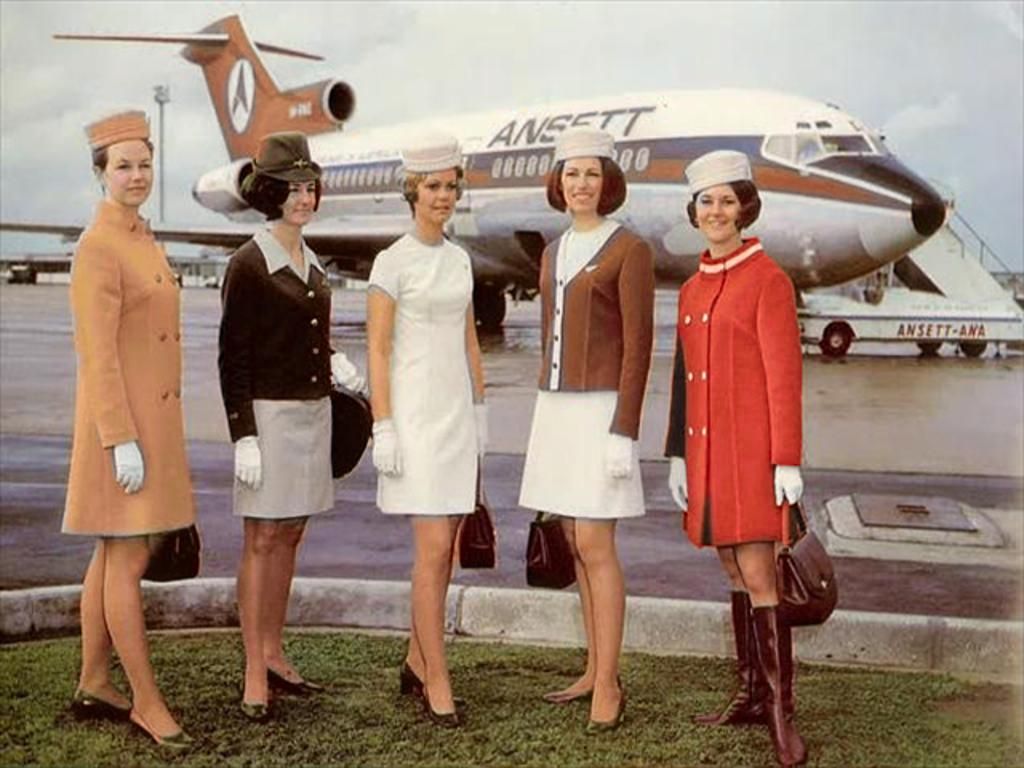}\\
        (5)&
\includegraphics[width=.18\textwidth]{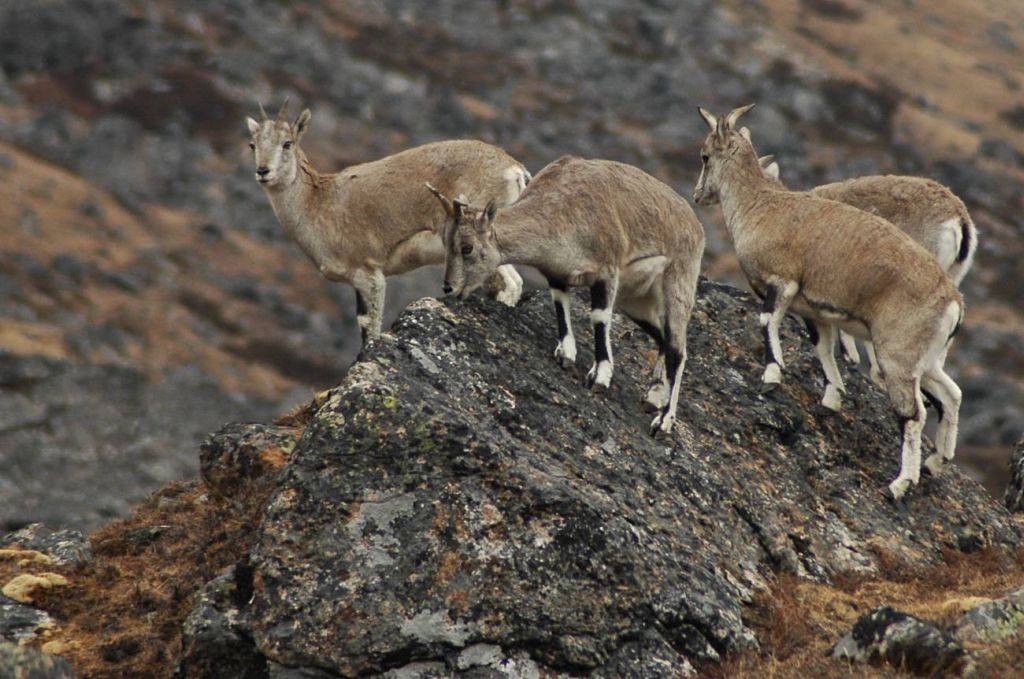}&
\includegraphics[width=.18\textwidth]{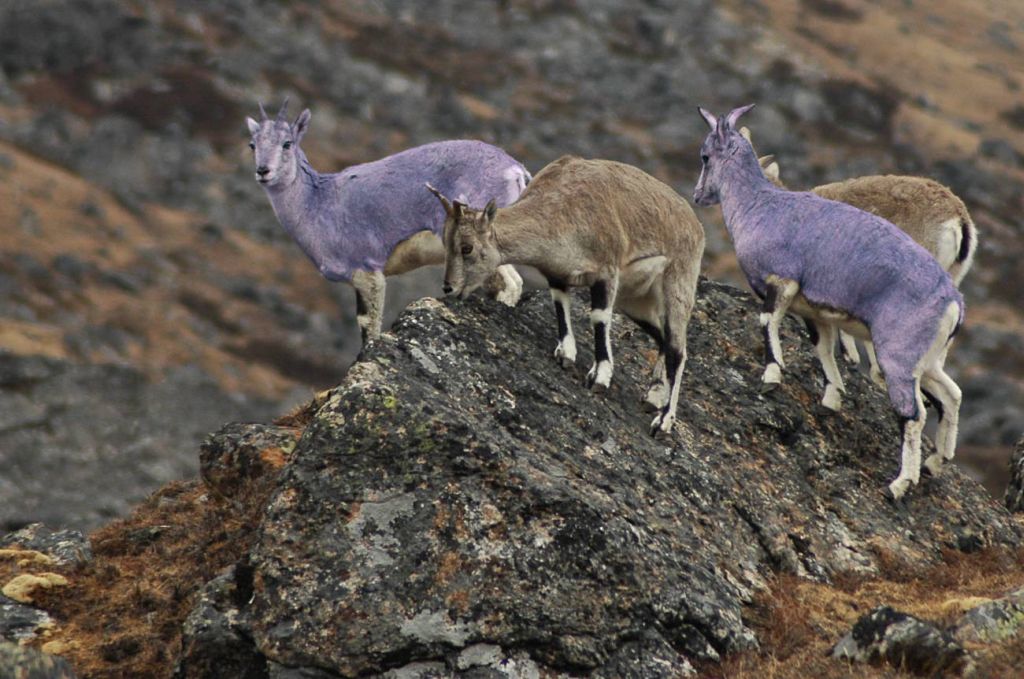}&
\includegraphics[width=.18\textwidth]{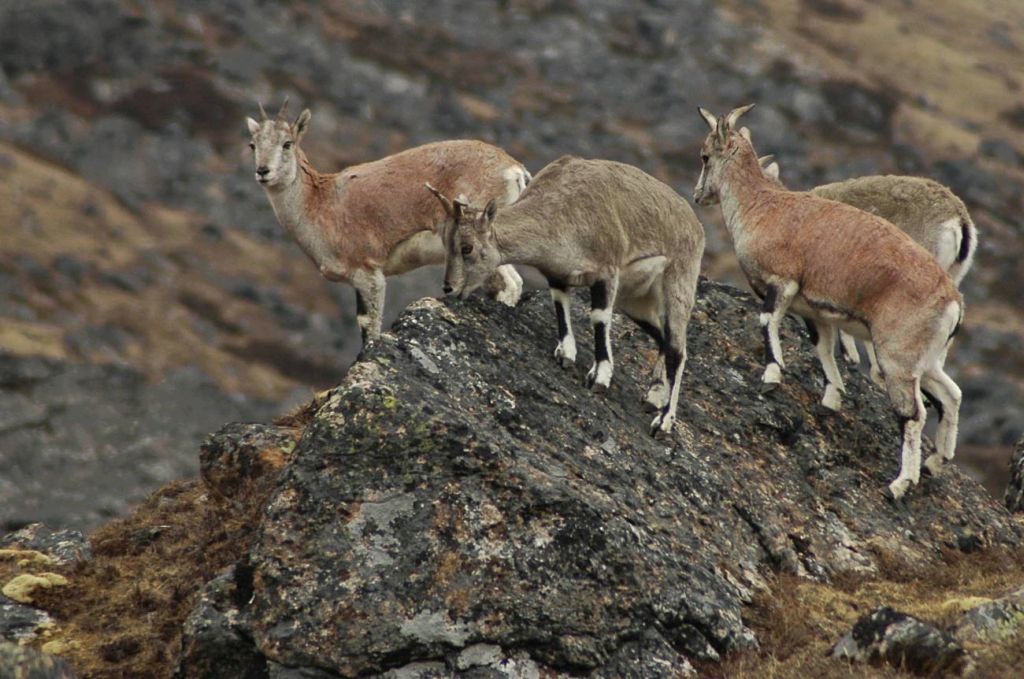}&
\includegraphics[width=.18\textwidth]{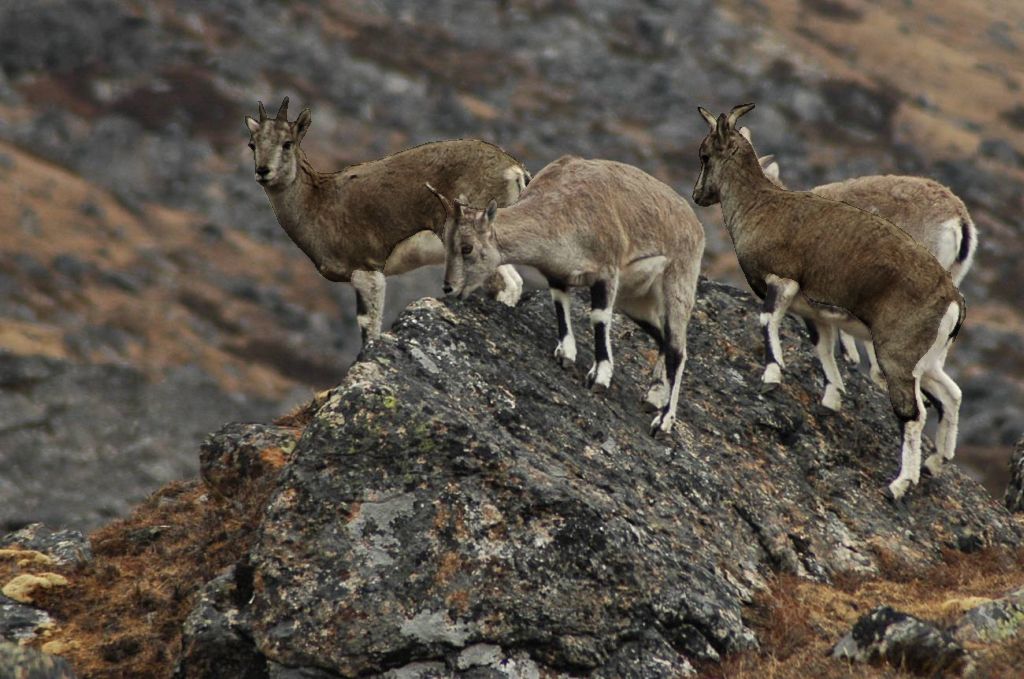}&
\includegraphics[width=.18\textwidth]{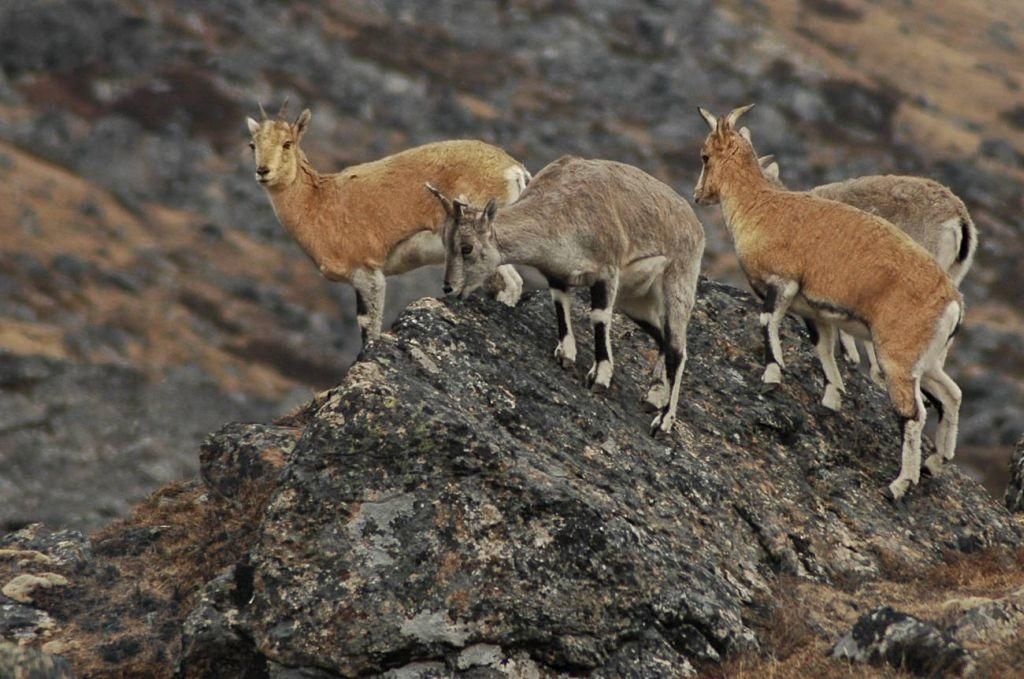}\\
        (6)&
\includegraphics[width=.18\textwidth]{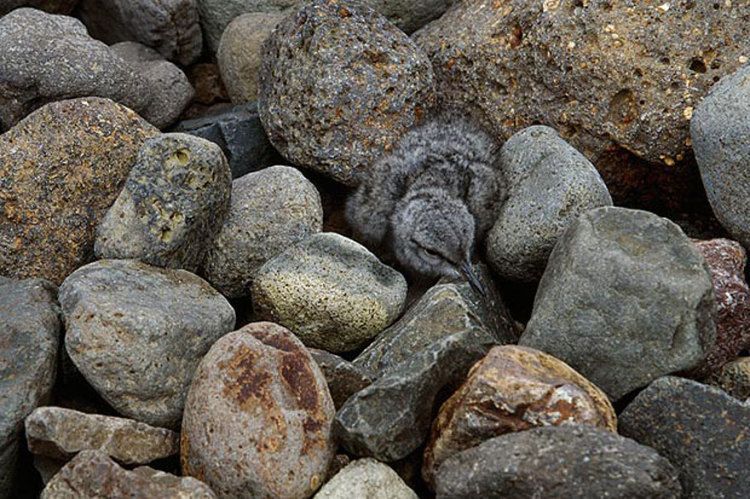}&
\includegraphics[width=.18\textwidth]{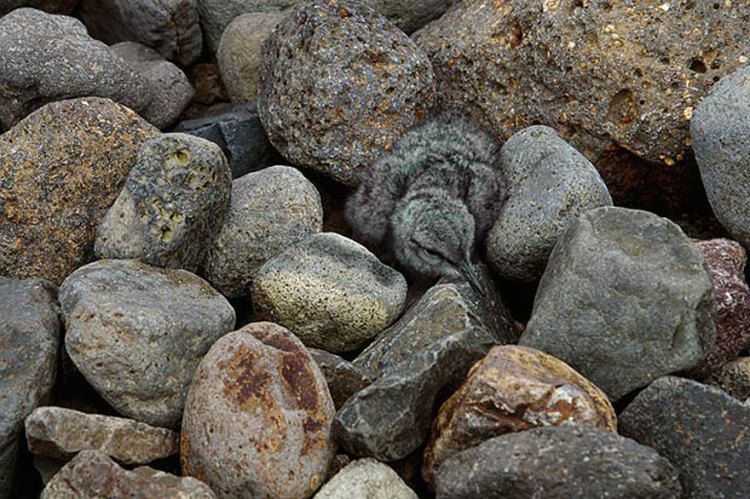}&
\includegraphics[width=.18\textwidth]{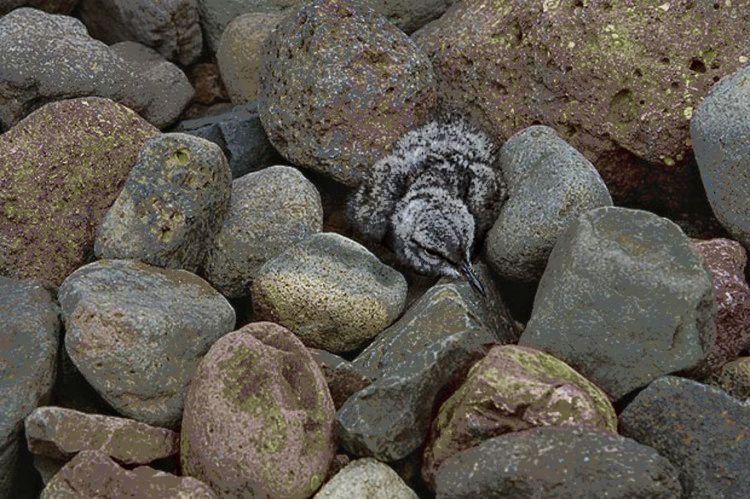}&
\includegraphics[width=.18\textwidth]{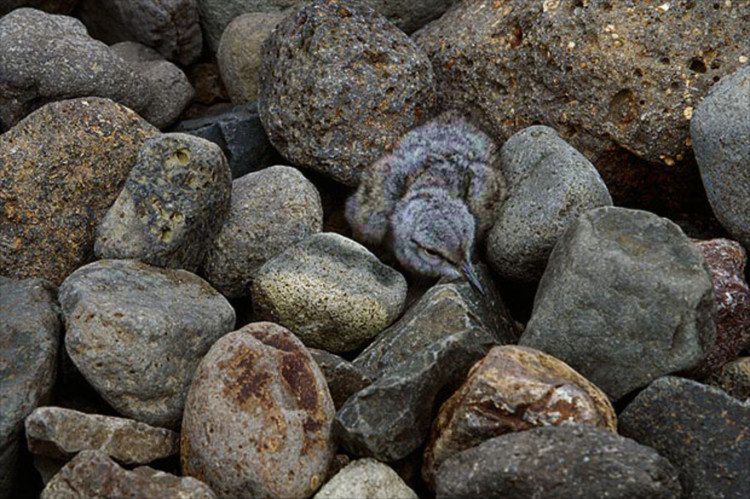}&
\includegraphics[width=.18\textwidth]{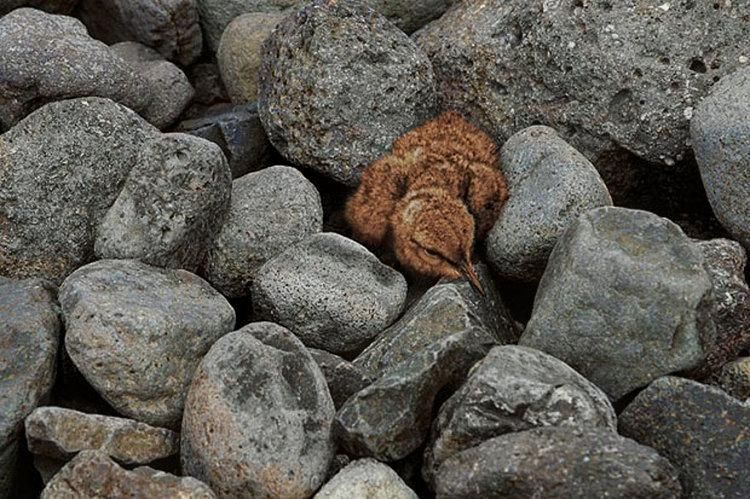}\\ 
        (7)&
        \includegraphics[width=.18\textwidth]{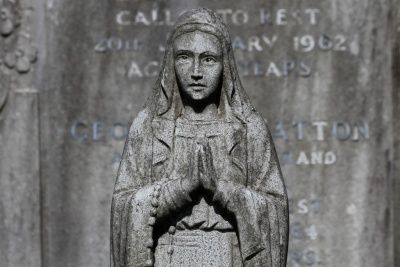}&
        \includegraphics[width=.18\textwidth]{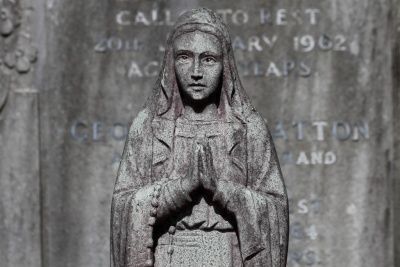}&
        \includegraphics[width=.18\textwidth]{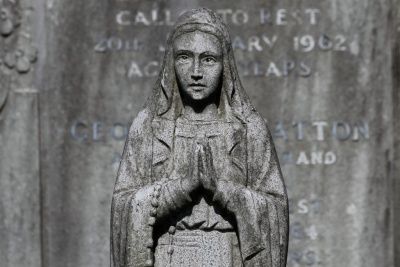}&
        \includegraphics[width=.18\textwidth]{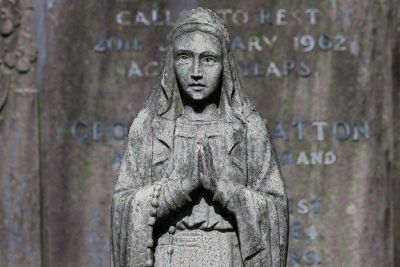}&
\includegraphics[width=.18\textwidth]{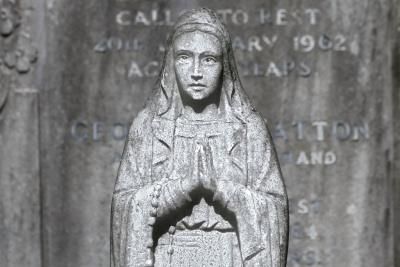}\\
& (a) Input image & (b) OHA & (c) HAG & (d) WSR & (e) Ours
    \end{tabular}
    \vspace{-0.2cm}
		\caption{\textbf{Object Enhancement} In these examples the user selected a target region to be enhanced (top row). To qualitatively assess the enhancement effect one should compare the input images in (a) to the manipulated images in (b,c,d,e), while considering the input mask (top). The results of OHA in (b) are often non realistic as they use arbitrary colors for enhancement. HAG (c) and WSR (d) produce realistic results, but sometimes (e.g., rows 1,2,6 and 7) they completely fail at enhancing the object and leave the image almost unchanged. Our manipulation, on the other hand, consistently succeeds in enhancement while maintaining realism. Our enhancement combines multiple saliency effects: emphasis by illumination (rows 1 and 7), emphasis by saturation (rows 2, 3 and 4) and emphasis by color (rows 1, 4-7).
}
\label{fig:enhance}
\end{figure*}

\section{Conclusions and Limitations}
\label{sec:conclusion}

We propose a general visual saliency retargeting framework that manipulates an image to achieve a saliency change, while providing the user control over the level of change. Our results outperform the state of the art in object enhancement, while maintaining realistic appearance. Our framework is also applicable to other image editing tasks such as distractors attenuation and background decluttering. Moreover, We establish a benchmark for measuring the effectiveness of algorithms for saliency manipulation.

Our method is not without limitations. First, since we rely on internal patch statistics, and do not augment the patch database with external images, the color transformations are limited to the color set of the image. 
Second, since our method is not provided with semantic information, in some cases the manipulated image may be non-realistic. For example, in Figure~\ref{fig:distractors}, the balloon is colored in gray, which is an unlikely color in that context.
Despite its limitations, our technique often produces visually appealing results that adhere to the user's wish.
\\

\noindent\textbf{Acknowledgements} This research was supported by the Israel Science Foundation under Grant 1089/16, by the Ollendorf foundation and by Adobe

\clearpage
{\small
\bibliographystyle{ieee}
\bibliography{paper_s}
}


\end{document}